\documentclass[lettersize,journal]{IEEEtran}
\usepackage{xcolor}
\usepackage{amsmath,amssymb,amsfonts}
\usepackage{algorithmic}
\usepackage{algorithm}
\usepackage{array}
\usepackage[caption=false,font=footnotesize,labelfont=rm,textfont=rm]{subfig}
\usepackage{textcomp}
\usepackage{stfloats}
\usepackage{url}
\usepackage{verbatim}
\usepackage{graphicx}
\usepackage{cite}
\usepackage{orcidlink}
\hypersetup{hidelinks}
\usepackage{wrapfig}
\usepackage{hyperref}
\usepackage{makecell}

\usepackage{enumitem}
\usepackage{tabularx}
\usepackage{adjustbox}
\usepackage{multirow}
\usepackage{multicol}
\usepackage{booktabs}
\usepackage{threeparttable}
\setlist[itemize]{leftmargin=*}

\begin{document}
\title{SMART: Scene-motion-aware human action recognition framework for mental disorder group}
\author{Zengyuan Lai$^{\orcidlink{0009-0005-3805-683X}}$, ~\IEEEmembership{Graduate Student Member,~IEEE}, Jiarui Yang$^{\orcidlink{0009-0005-6728-1489}}$, ~\IEEEmembership{Graduate Student Member,~IEEE}, Songpengcheng Xia$^{\orcidlink{0000-0002-9452-7967}}$, ~\IEEEmembership{Graduate Student Member, ~IEEE}, Qi Wu, Zhen Sun,\\ Wenxian Yu$^{\orcidlink{0000-0002-8741-776X}}$, ~\IEEEmembership{Senior Member, ~IEEE}, and Ling Pei$^{\orcidlink{0000-0003-1025-1260}}$, ~\IEEEmembership{Senior Member, ~IEEE}
\thanks{ Manuscript received xx June 2024; revised * 2024; accepted * 2024. Date of publication * 2024; date of current version * 2024. \textit{(Corresponding author: Ling Pei.)}}
\thanks{This work was supported by the National Natural Science Foundation of China (NSFC) under Grant 62273229.}
\thanks{The authors are with Shanghai Key Laboratory of Navigation and Location-based Services, School of Electronic Information and Electrical Engineering, Shanghai Jiao Tong University, Shanghai 200240, China (e-mail:zy.lai; jr.yang; songpengchengxia; robotics\_qi; zhensun; wxyu; ling.pei@sjtu.edu.cn).}}

\markboth{Journal of \LaTeX\ Class Files,~Vol.~14, No.~8, June~2024}%
{Zengyuan Lai \MakeLowercase{\textit{et al.}}: SMART: Scene-motion-aware human action recognition framework for mental disorder group}

\IEEEpubid{0000--0000/00\$00.00~\copyright~2024 IEEE}

\maketitle

\begin{abstract}
Patients with mental disorders often exhibit risky abnormal actions, such as climbing walls or hitting windows, necessitating intelligent video behavior monitoring for smart healthcare with the rising Internet of Things (IoT) technology. However, the development of vision-based Human Action Recognition (HAR) for these actions is hindered by the lack of specialized algorithms and datasets. In this paper, we innovatively propose to build a vision-based HAR dataset including abnormal actions often occurring in the mental disorder group and then introduce a novel Scene-Motion-aware Action Recognition Technology framework, named SMART, consisting of two technical modules. First, we propose a scene perception module to extract human motion trajectory and human-scene interaction features, which introduces additional scene information for a supplementary semantic representation of the above actions. Second, the multi-stage fusion module fuses the skeleton motion, motion trajectory, and human-scene interaction features, enhancing the semantic association between the skeleton motion and the above supplementary representation, thus generating a comprehensive representation with both human motion and scene information. The effectiveness of our proposed method has been validated on our self-collected HAR dataset (MentalHAD)
, achieving 94.9\% and 93.1\% accuracy in un-seen subjects and scenes and outperforming state-of-the-art approaches by 6.5\% and 13.2\%, respectively. The demonstrated subject- and scene- generalizability makes it possible for SMART's migration to practical deployment in smart healthcare systems for mental disorder patients in medical settings. The code and dataset will be released publicly for further research: \url{https://github.com/Inowlzy/SMART.git}.
\end{abstract}

\begin{IEEEkeywords}
Action Recognition, Mental Disorders, Multi-stage Fusion, Scene Semantic Understanding, Smart Healthcare
\end{IEEEkeywords}

\section{Introduction}
\label{introduction}
\begin{figure*}[htbp!]
\centering
\includegraphics[width=\linewidth]{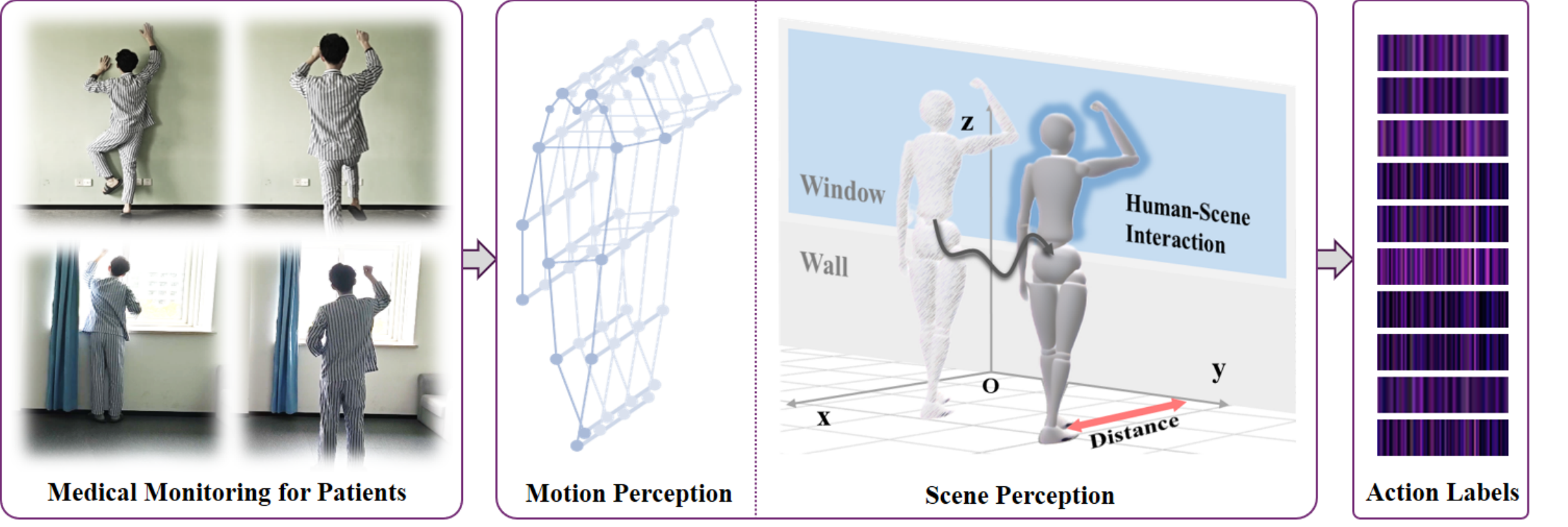}
\caption{The pipeline of scene-motion-aware HAR}
\label{intro}
\end{figure*}

\IEEEPARstart{P}{atients} with mental disorders may exhibit abnormal behaviors, such as climbing walls and hitting windows, significantly risky in medical institutions\cite{hao2021end}. With the advancements in IoT technology\cite{pei2021mars, hu2024end} and rising emphasis on medical healthcare\cite{tan2023quantitative, adil2024healthcare}, monitoring these actions intelligently draws significant attention. Traditional monitoring methods are time-consuming and labor-intensive. Deep learning-based Human Action Recognition (HAR) emerges as a promising alternative. Vision-based methods\cite{hu2024end,duan2022revisiting, ahmed2021deep} offer convenience, cost-effectiveness, intuitive visualization, and non-invasiveness compared to other modalities like the Inertial Measurement Units (IMUs)\cite{xia2024timestamp,pei2021mars} and millimeter-wave radars\cite{yang2024mmbat,waqar2023simulation}. Therefore, our work focuses on vision-based recognition of behaviors common in the mental disorder group, particularly abnormal behaviors. However, vision-based HAR primarily focuses on normal actions like daily actions\cite{wang2023videomae} and sports actions\cite{duan2022revisiting}, neglecting the above-mentioned abnormal actions that often involve interactions with surrounding scene elements, such as walls and windows. This limitation leads to significant challenges in implementing vision-based HAR methods for these abnormal actions.

The first challenge is data scarcity on abnormal actions\cite{chang2022contrastive}, significantly obstructing model training under the supervised approach. This scarcity arises from the infrequent occurrence of abnormal actions and the lack of cooperation from patients in medical institutions during data collection. Previous works on similar actions, such as fall-down in the elderly \cite{chen2023fall, wu2023video, zhang2023deep}, mainly acquire data through imitation by normal individuals due to actions' risks for the elderly, but they do not include the abnormal actions we focus on. To address this challenge, we construct a dataset by mimicking the abnormal actions in both daily scenes and wards. This leads to high demands on subject and scene generalizability for its practical application, introducing additional challenges.
\IEEEpubidadjcol

The second challenge concerns inter-person variability\cite{chen2019semisupervised,bulling2014tutorial}, which means different people can perform the same action differently. It demands our model to focus on the motion difference between activities, especially with the limited subjects in the training data. Skeleton motion features extracted from RGB videos have been proven to highlight dynamic changes in human poses and effectively preserve subject-independent information for HAR\cite{zhang2023dynamic,sun2022human}. Although skeleton data sacrifices human appearance and scene information, it boasts high computational efficiency, lower data volume requirements\cite{zeng2022inversionnet3d}, camera viewpoint independence, and robustness to lighting variations and background noise\cite{sun2022human}. Moreover, we can utilize a motion perception method to easily extract the skeleton motion feature, taking advantage of large amounts of skeleton datasets and robust pre-trained models. All the above allows the skeleton motion feature as our base representation of human motion to empower our model with strong subject generalizability for all actions.

The third challenge lies in inter-class similarity\cite{chen2019semisupervised,bulling2014tutorial}, where the similarity in human motions of different actions results in similar semantic representations and confusion. Particularly for abnormal actions like hitting a window versus hitting an object, the human poses are similar, making them difficult to distinguish using only skeleton motion features. To address this, considering these actions have different interactions between the human and scene elements, which means there are distinct spatial position differences between the human and scene elements, we need to obtain their spatial positions\cite{fiehler2023spatial} first and then calculate their difference. We propose using depth and semantic mask data to represent the spatial positions of humans and scene elements, obtained through pre-trained monocular depth estimators and semantic segmentation models. By further extracting the spatial position differences, we can represent human-scene interaction features, revealing interaction semantics such as a person close to a wall or window. Moreover, extracting a 3D motion trajectory of humans further enhances skeleton motion features, mitigating depth confusion. The above process, named scene perception, then helps distinguish abnormal actions.

To effectively integrate these features, we aim to design a multi-stage fusion network. In Stage I, motion-enhanced fusion, the motion trajectory is fused with the skeleton motion, forming a more comprehensive representation of human motion independent of scene elements. In Stage II, interaction-enhanced fusion, a scene weight network dynamically weighs the human-scene interaction for each action category. This stage determines the extent to which interaction features are used, as some actions may not involve interactions with scene elements. The weighted human-scene interaction feature is then fused with the human motion feature, creating a semantically complete representation of all actions. Our fusion method ensures consistent performance and strong scene generalizability across all actions. The schematic pipeline of our method is depicted in Fig. \ref{intro}.

The main contributions of our work are stated as follows:
\begin{itemize}
\item We present SMART, a scene-motion-aware HAR framework, which integrates human motion and human-scene interaction extracted from RGB videos. It generates a comprehensive semantic representation for accurately recognizing abnormal actions, especially for the mental disorder group.
\item For effectively utilizing additional scene information in videos, we propose a scene perception module, which extracts motion trajectory and human-scene interaction features from depth and semantic segmentation masks for supplementing skeleton motion representation.
\item To adequately enhance semantic associations among the extracted features, we design a multi-stage fusion network that contains motion-enhanced and interaction-enhanced fusion, significantly boosting subject and scene generalizability.
\item We construct a vision-based HAR dataset, MentalHAD, including six normal actions and four abnormal actions across three scenes and five subjects, totaling nearly 274 minutes. Our framework performs best compared with state-of-the-art methods and achieves strong subject and scene generalizability, facilitating its migration to the mental disorder group in medical settings for IoT-based smart healthcare.
\end{itemize}

The article proceeds as follows: Section II reviews vision-based HAR methods for abnormal actions of the mental disorder group and scene-aware HAR methods. Section III elaborates on the framework and its components. Section IV conducts and analyses comparison experiments, ablation studies, and clinical evaluation. Section V concludes our study.

\begin{figure*}
\includegraphics[width=\linewidth]{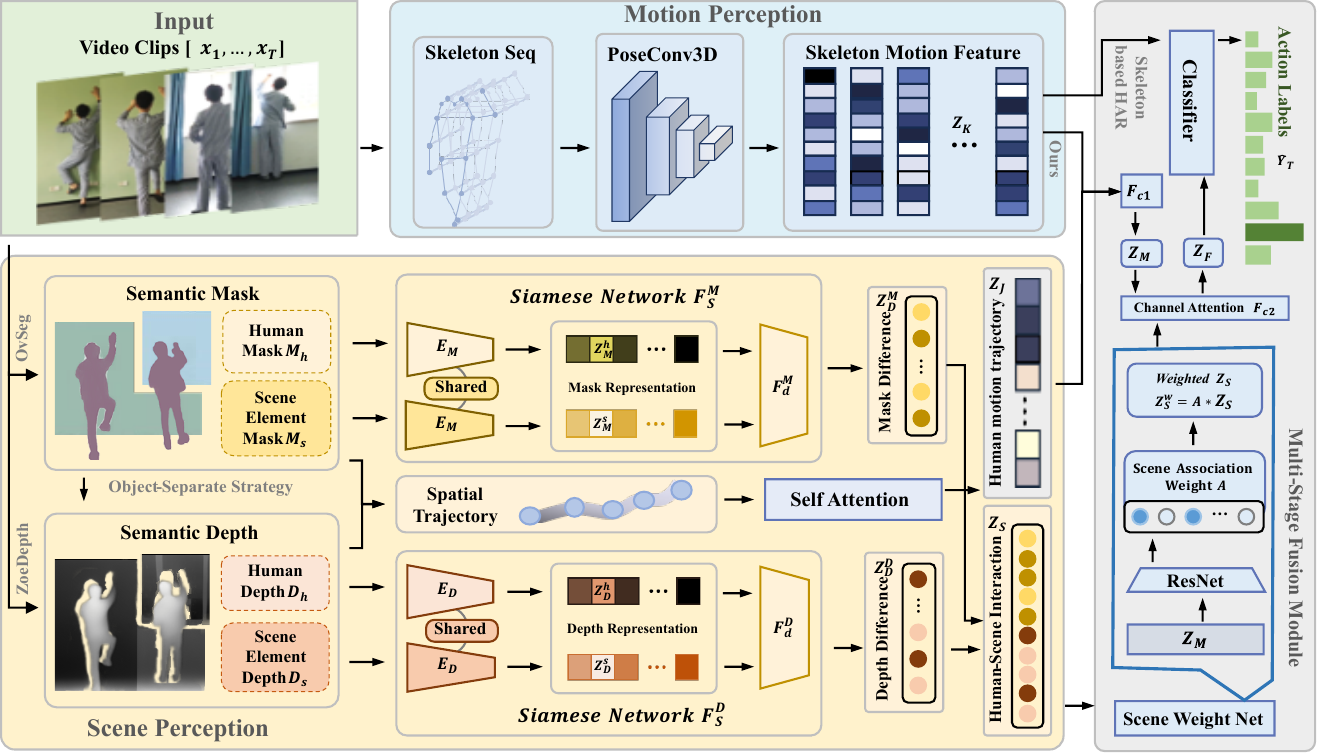}
\caption{Overview of SMART. (a) Shallow Information Decoupling Module (SIDM); (b) Motion Perception Module (MPM); (c) Scene Perception Module (SPM); (d) Feature Fusion and Classification Module (FFCM)}
\label{overall}
\end{figure*}

\section{Related Works}
\subsection{Vision-based HAR for mental disorder group}
Recent advancements in deep learning have revolutionized vision-based HAR to perform well on various actions\cite{wang2023videomae, duan2022revisiting}. According to the input data type, existing methods are divided into RGB-based and skeleton-based. The former has rich information but introduces redundancy and requires extensive data\cite{zeng2022inversionnet3d}. In contrast, the latter offers an efficient and robust alternative for HAR\cite{10230011, 10413307, duan2022revisiting,qiu2021skeleton}, where PoseC3D performs best on the NTU RGB+D 120 dataset\cite{duan2022revisiting}.

However, the abnormal action data scarcity we stated in the introduction hinders the vision-based HAR for the mental disorder group. From data aspects, Arifoglu et al.\cite{arifoglu2019detection} generate synthetic data on abnormal actions of dementia patients whose abnormal attributes exist in their occurring frequency, while Y. Hao et al.\cite{hao2021end} collect violent action segments related to mental disorders from the existing UCF-Crime dataset \cite{sultani2018real}. Nevertheless, abnormal actions often interact with scene elements in medical scenes, such as climbing walls and hitting windows, and are not involved. Others address this problem by introducing additional information. Chapron et al.\cite{chapron2019highly} utilize infrared proximity information for bathroom activity recognition. Moreover, Hao et al.\cite{9853267} explore incorporating emotional information to supplement skeleton motion data, demonstrating the effectiveness of additional information. However, the subjective nature of emotional data influences reliability. This limitation inspires us to explore more objective and practical supplementary information. In our work, we use scene information extracted from depth and semantic mask data to complement the base skeleton data.

\subsection{Scene-aware HAR}
Human action recognition (HAR) primarily focuses on skeleton motion features. However, for patients with mental disorders, these abnormal actions often involve interactions with scene elements, indicating the potential benefit of incorporating scene information into HAR.

Since Chen et al.\cite{6359337} pioneered integrating scene information in vision-based HAR, subsequent works\cite{6706382,6977062} extract and fuse scene information for HAR assistance, yielding promising results on benchmark datasets\cite{liu2009recognizing, reddy2013recognizing}. These works highlight the significance of robust feature extraction and fusion methods. Recent deep learning advancements for HAR use CNNs to encode static and dynamic scene elements \cite{7532632}. FASNet \cite{8101020} extracts action and scene features with a deep neural network. He et al. \cite{10006458} enhance extraction by integrating skeleton-based networks with scene interaction attributes. Zhai et al. \cite{10378184} introduce adaptive scene classification for scene-invariant information. To effectively utilize scene information, various fusion methods are employed. FASNet \cite{8101020} integrates features using a novel fusion and decision mechanism. He et al. \cite{10006458} improve fusion with channel attention mechanisms \cite{wang2020eca, 10007033} for adaptive feature weights. As He et al. suggest, selective integration of scene features prevents misrecognition. Inspired by all of the above, we use depth and semantic mask data to extract subject-invariant and scene-invariant features, enhancing robustness across actions.

\section{Methods}
This section introduces the proposed scene-motion-aware action recognition framework SMART, as shown in Fig. \ref{overall}. Our approach contains three modules: 1) a motion perception module to extract the skeleton motion feature as our baseline; 2) a scene perception module to extract human-scene interaction and human motion trajectory features as the supplement for action semantics; 3) a feature fusion and classification module to optimal feature utilization.

\subsection{Problem Statement}
In this article, we define the RGB sequence inputs as $X_{T}=[x_1,...,x_T],x_t \in R^{H \times W \times 3}$, and the corresponding sequence-level labels as $Y_T=[y_1,...,y_M],y_m \in R^C$, where $C$ is action classes, $T$ is the length of the RGB sequence, and $M$ is the total number of persons in one sequence. Traditional skeleton-based HAR methods utilize a pose estimation and feature extraction pipeline to obtain skeleton motion features as $Z_K=[z_K^1,...,z_K^M],z^m_K\in R^{N\times 2}$, where $N$ is the keypoints' total number of one person, with 2 dimensions as u, v coordinates. Based on skeleton motion features $Z_K$ and truth labels $Y_T$, the model is trained to predict action categories $\hat{Y}_K$.

In our work, instead of applying the skeleton motion feature $Z_K$ alone, our goal is to predict the sequence-level labels $\hat{Y}_T$ from the fusion of $Z_K$, the human motion trajectory feature $Z_J$, and the human-scene interaction feature $Z_S$, which will improve abnormal action recognition and enhance subject and scene generalizability. These additional features can be archived utilizing a scene perception network. Fusing them with the skeleton motion feature under a multi-stage fusion network, we can obtain the fused feature as $Z_F$. Then, with a classifier, our model could predict sequence-level labels $\hat{Y}_T$ from fused features $Z_F$.

\subsection{Motion Perception Module}
From previous works\cite{zeng2022inversionnet3d, duan2022revisiting}, skeleton-based HAR methods ensure robustness across subjects while preserving privacy by excluding identifiable features like faces. To ensure our subject generalizability, the skeleton motion feature $Z_K$ is extracted using a pre-trained model PoseC3D\cite{duan2022revisiting} $F_M$, shown in Fig. \ref{overall}, which contains three modules in order: human detection, 2D top-down pose estimation, and feature extraction.

When the skeleton motion feature is directly fed into a classifier $C_K$ to predict the action categories $\hat{Y}_K$, the full pipeline forms our baseline. With the available truth labels $Y_T$, we can optimize the parameters of our baseline by minimizing
\begin{equation}
\begin{aligned}
    L_K &= \sum\limits l(\hat{Y}_K, Y_T)\\
    &= \sum\limits l(C_K(F_M(X_{T}))), Y_T)
\end{aligned}
\end{equation}
where $l(\cdot)$ is the cross-entropy loss for classification.

However, for the baseline, there are still some shortcomings in solving the actual HAR problem for abnormal actions of the mental disorder group: the baseline ignores scene information implying the interaction relationship between humans and scene elements in RGB videos, thus causing poor performance on abnormal actions interacting with scene elements. Therefore, we introduce a scene perception module and a multi-stage fusion module to address these issues.

\subsection{Scene Perception Module}
\label{spm}
To effectively recognize abnormal actions with the assistance of scene information, we develop a scene perception module to extract scene information, from where we further extract human-scene interactions and motion trajectories. From section \ref{introduction}, human-scene interaction features $Z_S$ provide action semantics about how humans interact with scene elements, crucial for understanding abnormal behaviors. Meanwhile, human motion trajectory features $Z_J$ offer additional spatial dimensions beyond skeleton motion features, enhancing motion representation. To obtain them, we analyze spatial position differences between humans and scene elements for interaction features and obtain motion trajectory features from 3-D spatial positions involving both depth and image plane dimensions.

\begin{figure}
\includegraphics[width=\linewidth]{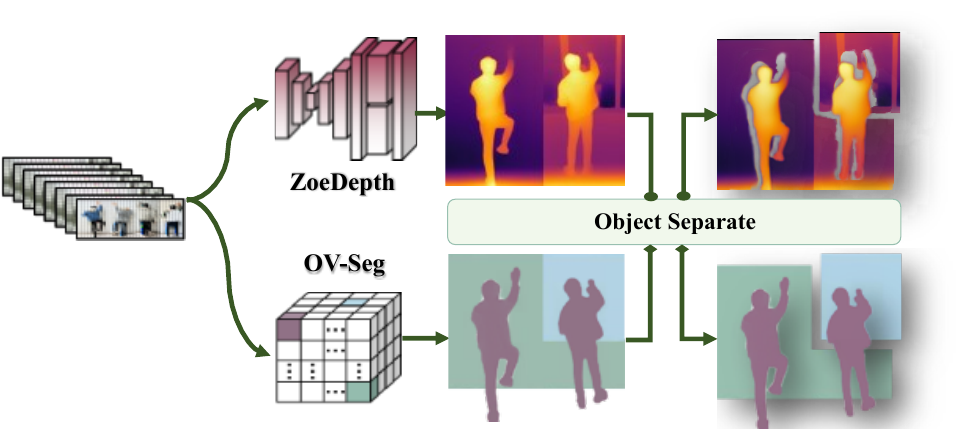}
\caption{Architecture of Depth and Semantic Mask Decoupling} \label{SIDM}
\end{figure}

To derive the mentioned spatial positions of humans and scene elements, we use depth and semantic mask data from monocular depth estimation and semantic segmentation. Depth data captures the distance of objects in each pixel from the camera in depth dimension, while semantic mask data defines the boundaries and categories of distinct regions in the image plane. Based on RGB sequence inputs \(X_{T}\), ZoeDepth\cite{https://doi.org/10.48550/arxiv.2302.12288} $F_D$ extracts the absolute distance of humans and scene elements as depth data, and OVSeg\cite{liang2023open} $F_M$ extracts categories and boundary coordinates as semantic mask data, shown in Fig. \ref{SIDM}. By separating the depth data using the object category of each pixel from semantic masks, we integrate class semantics into the depth data, generating semantic depth. The concatenation of semantic depth and masks forms the preliminary representation of spatial positions for each human and scene element. This allows us to explicitly calculate spatial position differences between humans and scene elements.

\begin{equation}
\begin{aligned}
    M_h, M_s = F_M(X_T),\quad
    D_h, D_s = F_D(X_T)
\end{aligned}
\end{equation}
where $D_h$, $D_s$, $M_h$, $M_s$ denote the depth of one human and scene element, the mask of one human and scene element. $F_M$ and $F_D$ are pre-trained models.

To further obtain the human-scene interaction feature $Z_S$, we use a Dual-Siamese Network\cite{sun2023siamohot} to extract spatial position differences between humans and scene elements from the provided preliminary representation of spatial positions. Each Siamese network extracts semantic depth or masks to generate a deep representation $Z_M^h, Z_M^s, Z_D^h, Z_D^s$ of spatial position for humans and scene elements in the depth dimension or image plane, respectively. Differences in these representations are then captured to represent interaction relationships accordingly, named human-scene interaction features $Z_S$.

\begin{equation}
\begin{aligned}
Z_M^h, Z_M^s &= E_M(M_h, M_s)\\
Z_D^h,Z_D^s &= E_D(D_h,D_s)\\
Z_D^D = F_d^D(Z_D^h,Z_D^s)&,\quad Z_D^M = F_d^M(Z_M^h, Z_M^s)\\
Z_S = [&Z_D^D, Z_D^M]
\end{aligned}
\end{equation}
where $E_D$ and $E_M$ respectively denote the feature extractor in two Siamese networks, each comprising two sub-networks of identical architecture with shared weights. $F_d^D$ and $F_d^M$ are difference extractors of spatial position's deep representation. $Z_D^D$ and $Z_D^M$ denote spatial position differences in the depth dimension and image plane respectively.

To capture the human motion trajectory feature $Z_J$, we first construct a statistical center based on the human body's average depth and pixel coordinates in masks. Then, the motion trajectory $J$, shown in Fig. \ref{intro}, is constructed from the human body center. To extract the motion features in the trajectory, we leverage methods from time series analysis, treating the trajectory as a three-dimensional coordinate time series. The self-attention mechanism\cite{10007033} with relative position encoding stands out for its ability to dynamically focus on information across different positions in the sequence. Utilizing the self-attention network, the human motion trajectory feature $Z_J$ is extracted as follows.

\begin{equation}
\begin{aligned}
{Z_J} = Sof&tmax({\frac{QK^T+S_{rel}}{\sqrt{D_h}})V}\\
&S_{rel} = QR^T\\
Q = W^qJ&,K = W^kJ,V = W^vJ
\end{aligned}
\end{equation}
where ${Q}$, ${K}$, ${V}$ are the query matrix, key matrix, and value matrix, $D_h$ is the hidden dimension, ${R}$ is the relative position encoding matrix, ${S_{rel}}$ is the matrix with relative positions.

In the end, we utilize the scene perception module to extract the human-scene interaction feature $Z_S$ and the motion trajectory feature $Z_J$ as a supplement of action semantics representation beneficial for abnormal action recognition of the mental disorder group.

\subsection{Multi-Stage Fusion Module}
Considering the semantics association among the above features, we design a multi-stage fusion network to fuse the most associated features in the semantics stage by stage. In Stage I, motion-enhanced fusion, to get enhanced human motion representation $Z_M$, we fuse the skeleton motion feature $Z_K$ with the motion trajectory feature $Z_J$ using channel attention $F_{c1}$. The two features are assigned to separate channels to learn their respective contributions to human motion, thereby attaining the optimal representation of enhanced human motion features $Z_M= F_{c1}(Z_K, Z_J)$.

In Stage II, considering some actions have no interaction with scenes, we propose a scene weight network $F_w$ with a CNN architecture\cite{10007033} to dynamically weigh the human-scene interaction feature $Z_S$ based on various human-scene associations $A$ extracted from $Z_M$. In our network, this is supervised by human-scene association labels $Y^S_T$, with a value of 1 for actions interacting with scenes and 0 for others.
\begin{equation}
\begin{aligned}
A = F_w(Z_M),\quad
{Z_S^w} = A \cdot Z_S
\end{aligned}
\end{equation}

Then, to supplement the lack of interaction semantics in human motion, the weighted human-scene interaction feature $Z_S^w$ can be fused with the human motion feature $Z_M$ from stage I, with another channel attention. This stage finally generates a semantic-complete representation $Z_F$ of all actions. The fused feature $Z_F$ is then fed into a classifier $C_{K+S}$, a fully connected network, to predict action categories $\hat{Y}_T$. 

In the end, based on action category labels $Y_T$ and human-scene association labels $Y^S_T$, we can optimize our framework parameters by minimizing:
\begin{equation}
\begin{aligned}
    L_{K+S} &= \sum\limits l(\hat{Y}_T, Y_T)+l^B(A, Y^S_T)\\
    &= \sum\limits l(C_{K+S}(Z_F)), Y_T)+l^B(F_w(Z_M), Y^S_T)
\end{aligned}
\end{equation}
where $l(\cdot)$ is the cross-entropy loss for the classification and $l^B(\cdot)$ denotes the binary cross-entropy loss.

\section{Experiments and Discussion}
This section introduces our self-collected dataset: MentalHAD, recognition metrics, and experimental methodologies used in our experiments. Then, we evaluate the comprehensive performance and generalization ability across subjects and scenes, by comparing SMART with the state-of-the-art algorithm of vision-based HAR. The generalizability reveals the migration possibility from normal people to the mental disorder group and from normal scenes to medical settings. We also verify the effectiveness of each module through various quantitative and qualitative experiments. Finally, we conduct practical deployment for future application evaluation.

For the experimental setup, we implement our model on the Pytorch platform with an RTX 3090 GPU. Adam optimizer with a 0.001 learning rate is used in training, decaying by 0.0005 with patience step 5, and batch size is set up to 16.
\subsection{Evaluation Protocols and Dataset}
\label{protocols}
In our experiment, we quantitatively evaluate our model from four aspects of HAR on our dataset: MentalHAD. We employ the top-1 accuracy ($Acc$), macro precision ($P_m$), macro-recall ($R_m$), and macro F1-score ($F_m$) as the HAR metrics. Macro means the class-average method. Details of the MentalHAD are described in section \ref{dataset}.
\subsubsection{Evaluation Protocols}
In multi-class classification evaluation, accuracy has some limitations. The Macro-Average method calculates each category's average Precision, Recall, and F1 Score, ensuring equal treatment of all classes. In the following equations, $N$ denotes the class number, and $TP_i$, $FP_i$, $TN_i$, and $FN_i$ are the true positives, false positives, true negatives, and false negatives of class $i$, respectively.

\textbf{Accuracy}: $Acc$ is the ratio of correctly predicted samples across all classes.
\begin{equation}
    Acc = \frac{\sum_{i=1}^{N} (TP_i + TN_i)}{\sum_{i=1}^{N} (TP_i + TN_i + FP_i + FN_i)}
\end{equation}

\textbf{Macro-Precision}: $P_m$ indicates the proportion of true positive predictions in the positive class.
\begin{equation}
P_m = \frac{1}{N} \sum_{i=1}^{N} \frac{TP_i}{TP_i + FP_i}
\end{equation}

\textbf{Macro-Recall}: $R_m$ reflects the proportion of actual positive class samples correctly predicted as positive.
\begin{equation}
R_m = \frac{1}{N} \sum_{i=1}^{N} \frac{TP_i}{TP_i + FN_i}
\end{equation}

\textbf{Macro-F1 Score}: $F_m$ represents the harmonic mean of Precision and Recall, providing a balance between the two.
\begin{equation}
F_m = \frac{1}{N} \sum_{i=1}^{N} 2 \times \frac{P_m \times R_m}{P_m + R_m}
\end{equation}

To evaluate the overall performance, subject generalizability, and scene generalizability of SMART, we compare SMART with other state-of-the-art methods on MentalHAD, calculating all four metrics for all actions and three metrics except accuracy for abnormal actions. To verify the effectiveness of each module, we further conduct the ablation study on its overall performance and corresponding generalizability, evaluating the metrics $Acc$ and $F_m$ for all actions.

\subsubsection{Dataset}
\label{dataset}
Due to the difficulty in collecting abnormal action data performed by the mental disorder group, existing HAR datasets rarely include abnormal actions of such patients interacting with large-scale scene elements that we care about. To ensure the generalization ability of our framework and comprehensively evaluate its performance on such abnormal actions and normal actions, we select five normal persons to mimic\cite{shen2023indoor, wu2023video, zhang2023deep} climbing walls, hitting windows, climbing, and hitting, representing the above abnormal actions. 

\begin{figure}[t]
\centering
\includegraphics[width=\linewidth]{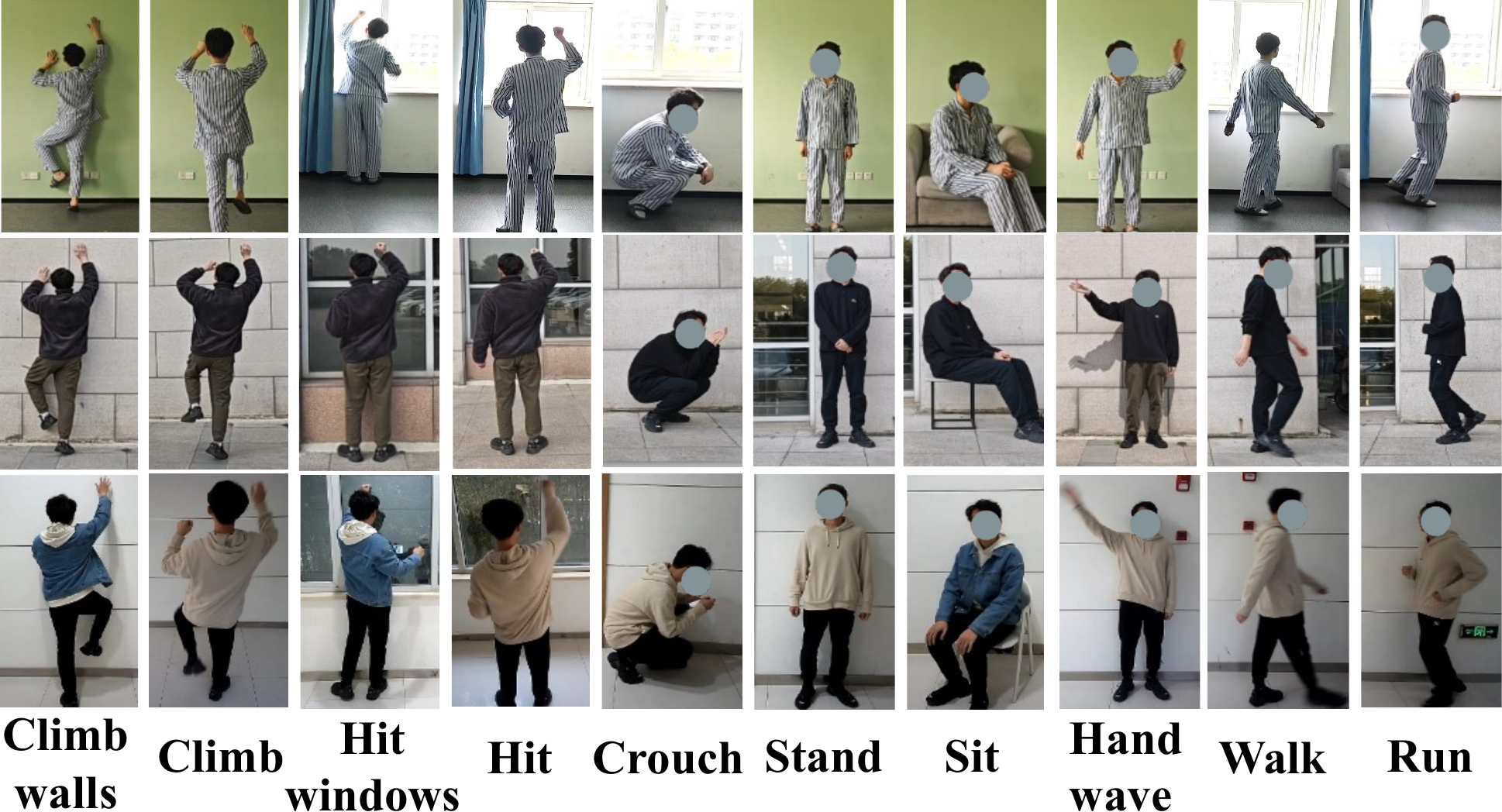}
\caption{Samples of MentalHAD in three scenes on subject A. (a) Bottom: scene I; (b) Medium: scene II; (c) Top: scene III.}
\label{datasetf}
\end{figure}

Collecting data with a HIKVISION USB Camera DS-E11, we build a dataset called MentalHAD with four abnormal actions (climbing walls, hitting windows, climbing, and hitting) and six normal actions (crouching, standing, sitting, hand waving, walking, and running). It includes RGB videos of about 274 minutes (493504 frames) with 30 FPS in three different scenes, five subjects, and seven scene-subject pairs. They are organized into 69 sequences, each containing the data of only one action in about 2-5 minutes. Samples of MentalHAD in all scenes on subject A are shown in Fig. \ref{datasetf}.

To meet the experimental needs for data, we split the MentalHAD dataset into the training set and testing set, specializing in model training and overall performance evaluation, shown in Table \ref{split}. The training set is split into three parts with a ratio of 7:2:1 for model training, validation, and internal testing. The testing set is an external testing set isolated from the training set, with new subjects and scenes, created for the overall performance evaluation of practical applications. It is further divided into two subsets to evaluate their contributions to improving subject or scene generalizability respectively. Videos of subject A in scene III are specially collected in a ward-like scene to mimic the practical scene\cite{song2024learning}, shown in the top line of Fig. \ref{datasetf}. Moreover, in the ablation study, different modules contribute differently to the two generalizabilities, which will be declared in the corresponding section. 

\begin{table}[t]
\Large
\renewcommand\arraystretch{1.5}
\centering
\caption{MentalHAD Dataset Split Strategy}
\label{split}
\begin{threeparttable}
\resizebox{\linewidth}{!}{
\begin{tabular}{c|c|c|c|c|c}
\toprule[1.1pt]
\multicolumn{2}{c|}{\textbf{Dataset Split}} & \textbf{Subjects} & \textbf{Scenes} & \multicolumn{2}{c}{\textbf{Settings}} \\
\midrule
\multicolumn{2}{c|}{\makecell[c]{Training Set}} & A, B, C & I & \multicolumn{2}{c}{\makecell[c]{Model training, validation,\\and internal testing with\\sequences 1-30\\random split by ratio of 7:2:1}} \\ \midrule
\multirow{3}*{{\makecell[c]{Testing\\Set}}}&
\makecell[c]{Testing\\Subset 1} & D & I & \makecell[c]{Setting I:\tnote{1}\\with sequences\\31-40} &\multirow{3}*{{\makecell[c]{Overall\\Evaluation\\with sequences\\31-69}}}\\ \cmidrule(l{1pt}r{1pt}){2-5}
&\multirow{2}*{{\makecell[c]{Testing\\Subset 2}}} & A, E& II & \multirow{2}*{\makecell[c]{Setting II:\tnote{2}\\with sequences\\41-69}} &\\ \cmidrule(l{1pt}r{1pt}){3-4}
&& A &III &&\\
\bottomrule[1.1pt]
\end{tabular}}
\begin{tablenotes}
  \footnotesize
  \item[1] Setting I: Subject generalizability evaluation;
  \item[2] Setting II: Scene generalizability evaluation;
  \item[3] Sequences 1-30 have 204288 frames, sequences 31-40 have 81728\\frames, and sequences 41-69 have 207488 frames.
  \end{tablenotes}
  \end{threeparttable}
\end{table}

Then, the following experiments can be divided into overall evaluation, Setting I test, and Setting II test, respectively, on the Testing Set, Testing Subset 1, and Testing Subset 2.

\subsubsection{Compared methods}
For overall performance evaluation, we compare our method with state-of-the-art methods in vision-based HAR, divided into RGB-based and skeleton-based methods based on the data used by the feature extraction module. We choose TimeSformer\cite{bertasius2021space}, VideoSwin\cite{liu2022video}, MViT V2\cite{li2022mvitv2} and Uniformer V2\cite{li2023uniformerv2} to represent the RGB-based methods, and select ST-GCN\cite{yan2018spatial}, 2s-AGCN\cite{shi2019two}, ST-GCN++\cite{duan2022pyskl}, and PoseC3D\cite{duan2022revisiting} to represent skeleton-based methods due to their proven performance, reproducibility, and available implementations. The methods are listed as follows:

\begin{table*}[htbp!]
\renewcommand\arraystretch{1}
\centering
\caption{Comparison Experiment}
\label{comparison}
\resizebox{\linewidth}{!}{
\begin{tabular}{c|cccc|cccc|cccc}
\Xhline{1pt}
\multicolumn{13}{c}{\textbf{All Actions}} \\
\toprule[1pt]
\multirow{2}*{\textbf{\raisebox{-6pt}{\makecell[c]{Method}}}}
&\multicolumn{4}{c}{\textbf{Overall}} \vline & \multicolumn{4}{c}{\textbf{Setting I}} \vline& \multicolumn{4}{c}{\textbf{Setting II}}\\ \cmidrule(l{1pt}r{0pt}){2-13}
&\textbf{$Acc$} & \textbf{\makecell[c]{$P_m$}} & \textbf{\makecell[c]{$R_m$}} & \textbf{\makecell[c]{$F_m$}} & \textbf{$Acc$} & \textbf{\makecell[c]{$P_m$}} & \textbf{\makecell[c]{$R_m$}} & \textbf{\makecell[c]{$F_m$}}& \textbf{$Acc$} & \textbf{\makecell[c]{$P_m$}} & \textbf{\makecell[c]{$R_m$}} & \textbf{\makecell[c]{$F_m$}}\\
\midrule
TimeSformer\cite{bertasius2021space} & 25.6\% &40.0\% & 25.9\%& 27.1\%& 49.0\% &46.2\% & 50.3\%& 41.8\%& 16.4\% &19.9\% & 15.7\%& 11.8\%\\
VideoSwin\cite{liu2022video} & 55.8\% & 57.3\% & 55.7\% & 55.3\% & 62.7\% & 67.0\% & 64.8\% & 62.2\% & 53.1\% & 52.3\% & 51.0\% & 49.1\% \\
MViT V2\cite{li2022mvitv2} & 30.1\% &41.0\% & 30.9\%& 31.2\%& 46.9\% &32.4\% & 50.2\%& 38.8\%& 27.1\% &40.0\% & 19.0\%& 17.1\%\\
Uniformer V2\cite{li2023uniformerv2} & 63.6\% &81.1\% & 63.5\%& 67.8\%& 76.8\% &78.6\% & 78.4\%& 78.4\%& 58.4\% &80.3\% & 56.4\%& 57.8\%\\ \midrule 
ST-GCN\cite{yan2018spatial} & 71.5\% &  82.8\% &  70.9\% & 72.2\% & 84.9\% & 81.6\% & 85.8\% & 82.9\% &66.2\% & 73.2\% & 63.4\% & 61.2\% \\
2s-AGCN\cite{shi2019two} & 72.9\% & 82.1\% & 72.3\% & 72.0\% & 88.0\% & 83.9\% & 88.8\% & 85.6\% & 67.0\% & 60.4\% & 63.9\% & 57.8\% \\
ST-GCN++\cite{duan2022pyskl} & 69.7\% & 73.5\% & 68.6\% & 68.1\% & 74.7\% & 67.8\% & 76.4\% & 70.4\% & 62.8\% & 60.5\% & 59.8\% & 56.3\%\\
PoseC3D\cite{duan2022revisiting} & {82.3\%} & {85.5\%} & {80.7\%} & {81.3\%} & {88.4\%} & {91.9\%} & {89.2\%} & {88.7\%} & {79.9\%} & {84.5\%} & {77.7\%} & {78.2\%} \\ \midrule
\textbf{Ours} & \textbf{93.6\% }& \textbf{93.9\%} &\textbf{92.8\%} & \textbf{93.1\%} & \textbf{94.9\%} & \textbf{95.6\%} & \textbf{95.2\%} & \textbf{95.1\%} & \textbf{93.1\%} & \textbf{93.6\%} & \textbf{92.5\%} & \textbf{92.6\%} \\ \midrule
\text{Gain} & +11.3\% & +8.4\% & +12.1\% & +11.8\% & +6.5\% & +3.7\% & +6.0\% & +6.4\% & +13.2\% & +9.1\% & +14.8\% & +14.4\% \\
\bottomrule[1pt]
\end{tabular}}
\resizebox{1\linewidth}{!}{
\begin{tabular}{c|>{\centering}p{1.2cm}p{1.2cm}p{1.2cm}|>{\centering}p{1.2cm}p{1.2cm}p{1.2cm}|>{\centering}p{1.2cm}p{1.2cm}p{1.2cm}}
\multicolumn{10}{c}{\textbf{Abnormal Actions}} \\
\toprule[1pt]
\multirow{2}*{\textbf{\raisebox{-6pt}{\makecell[c]{Method}}}}
&\multicolumn{3}{c}{\textbf{Overall}} \vline & \multicolumn{3}{c}{\textbf{Setting I}} \vline& \multicolumn{3}{c}{\textbf{Setting II}}\\ \cmidrule(l{1pt}r{0pt}){2-10}
&\textbf{\makecell[c]{$P_m$}} & \textbf{\makecell[c]{$R_m$}} & \textbf{\makecell[c]{$F_m$}} &\textbf{\makecell[c]{$P_m$}} & \textbf{\makecell[c]{$R_m$}} & \textbf{\makecell[c]{$F_m$}}& \textbf{\makecell[c]{$P_m$}} & \textbf{\makecell[c]{$R_m$}} & \textbf{\makecell[c]{$F_m$}}\\
\midrule
TimSformer\cite{bertasius2021space} & 35.3\% &34.7\% & 31.9\%& 41.6\%& 47.5\% &41.3\% & 16.9\%& 24.8\%& 15.7\%\\
VideoSwin\cite{liu2022video}&32.2\% & 35.4\% & 33.2\% & 33.6\% & 30.2\% & 26.8\% & 24.7\% & 32.7\% & 27.1\% \\
MViT V2\cite{li2022mvitv2} &26.5\% &35.2\% & 26.6\%& 26.2\%& 27.8\% &27.0\% & 31.7\%& 37.4\%& 28.2\%\\
Uniformer V2\cite{li2023uniformerv2} & 54.2\% &38.3\% & 38.6\%& 49.8\%& 49.6\% &49.7\% & 50.7\%& 29.1\%& 20.2\%\\ \midrule
ST-GCN\cite{yan2018spatial}&65.8\% & 47.3\% & 45.9\% & 58.6\% & 68.8\% & 61.9\% & 43.6\% & 34.1\% & 23.1\% \\
2s-AGCN\cite{shi2019two}&{69.4\%} & 49.8\% & 50.4\% & 61.8\% & 72.1\% & 65.1\% & 19.2\% & 36.2\% & 24.8\% \\
ST-GCN++\cite{duan2022pyskl}&65.3\% & 50.3\% & 53.7\% & 62.2\% & 69.3\% & 64.7\% & 42.9\% & 37.3\% & 36.6\% \\
PoseC3D\cite{duan2022revisiting}&67.5\% & {61.4\%} & {60.6\%} & {82.3\%} & {75.1\%} & {74.1\%} & {65.4\%} & {55.4\%} & {54.1\%} \\ \midrule
\textbf{Ours} &\textbf{92.1\%} & \textbf{91.0\%} & \textbf{91.4\%} & \textbf{90.7\%} & \textbf{89.2\%} & \textbf{89.2\%} & \textbf{92.9\%} & \textbf{92.3\%} & \textbf{92.6\%} \\ \midrule
\text{Gain} &  +22.7\% & +29.6\% & +30.8\% & +8.4\% & +14.1\% & +15.1\% & +27.5\% & +36.9\% & +38.5\% \\
\bottomrule[1pt]
\end{tabular}}
\end{table*}

\begin{figure*}[htbp!]
\centering
\subfloat[Uniformer V2]{\label{cuniformer}%
\includegraphics[width=0.327\linewidth]{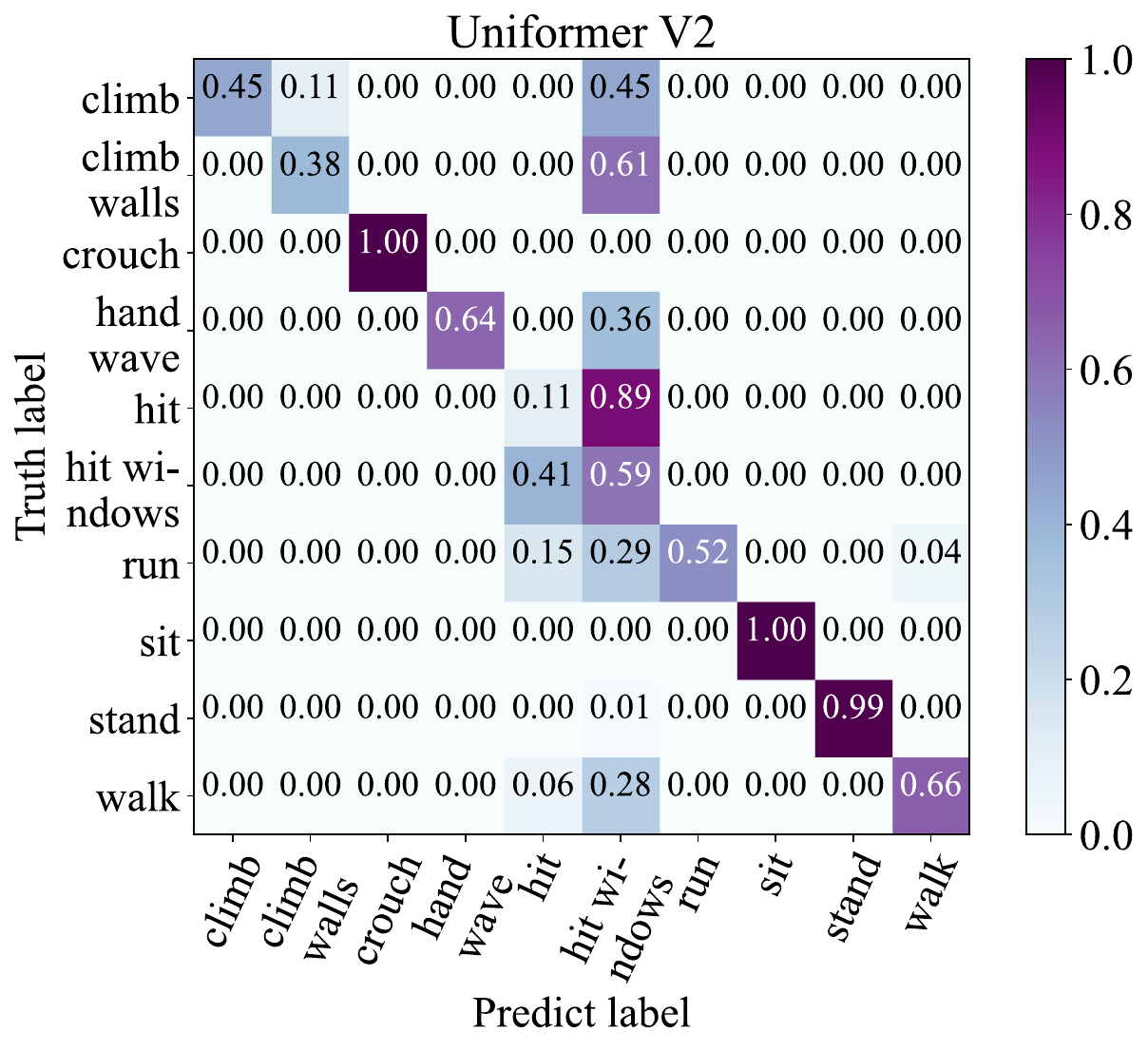}}
\hfil
\subfloat[PoseC3D]{\label{cp3d}%
\includegraphics[width=0.327\linewidth]{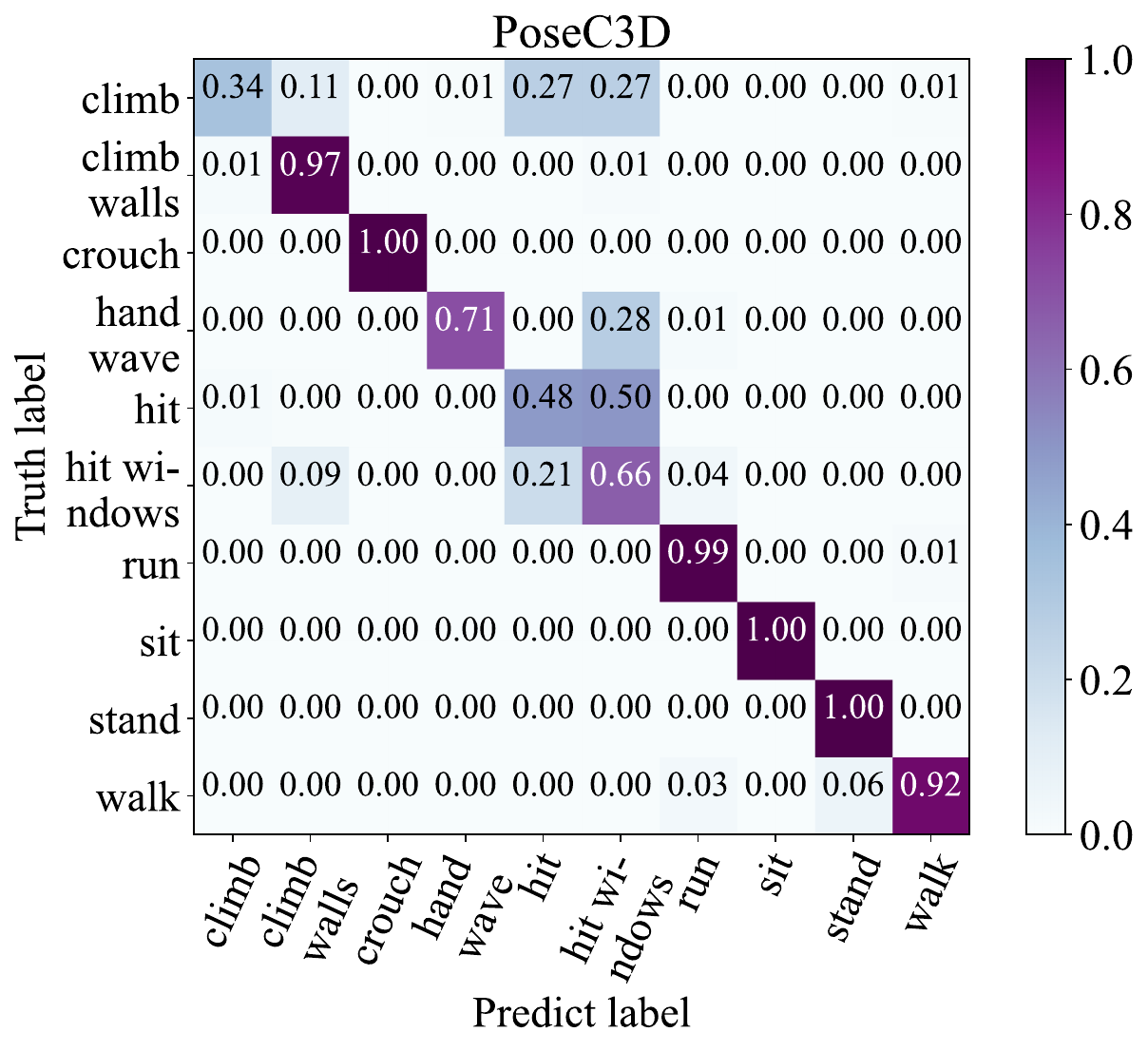}}
\hfil
\subfloat[Ours]{\label{cours}%
\includegraphics[width=0.327\linewidth]{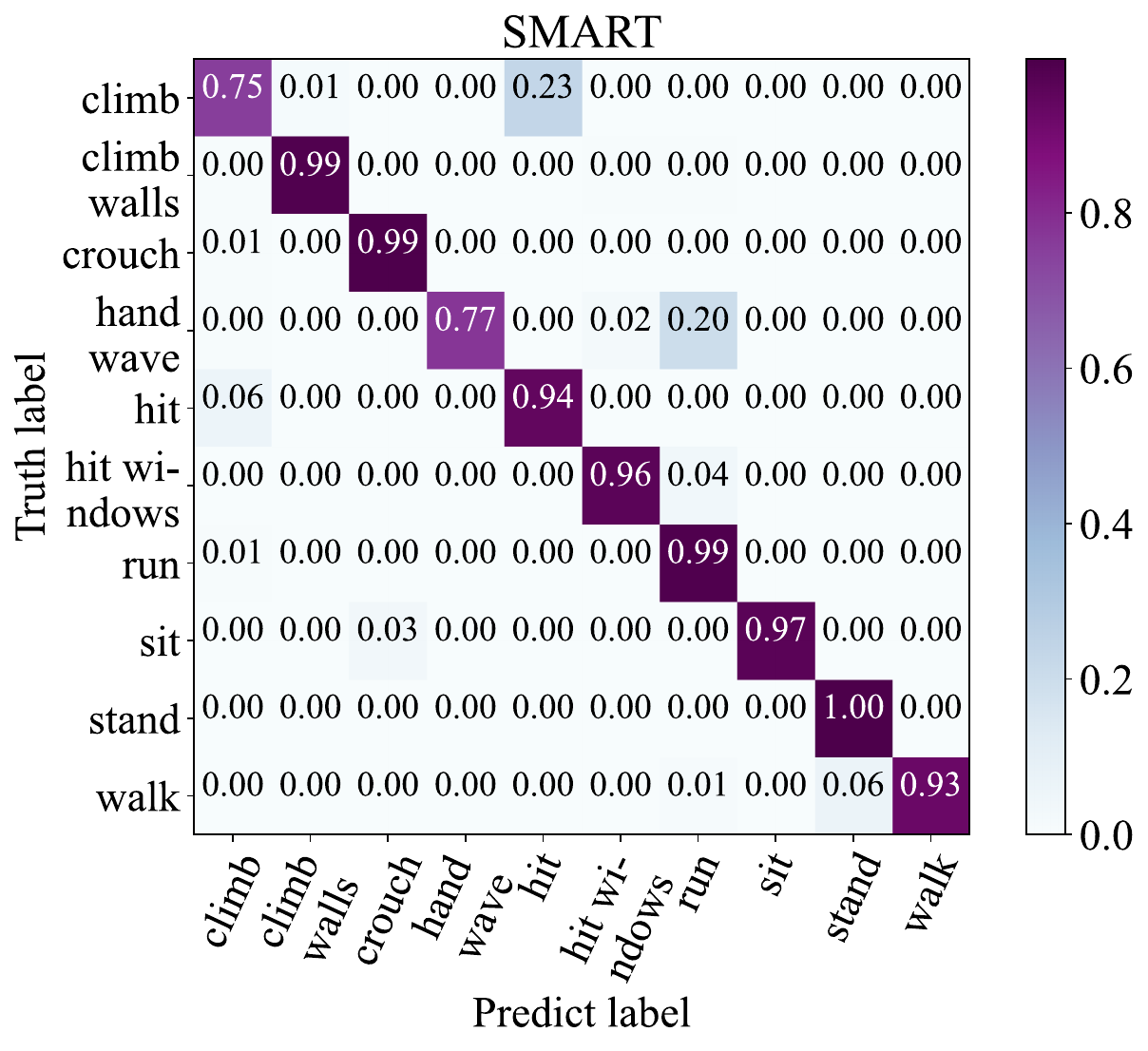}}
\caption{The confusion matrix of different methods on MentalHAD testing set. (a) Uniformer V2. (b) PoseC3D. (c) Ours.}
\label{conf}
\end{figure*}

\begin{itemize}
    \item \textbf{TimeSformer\cite{bertasius2021space}:} A Transformer-based model designed for effective temporal information capture in videos.
    \item \textbf{VideoSwin\cite{liu2022video}:} A Swin Transformer architecture to capture spatial-temporal features from RGB videos for HAR.
    \item \textbf{MViT V2\cite{li2022mvitv2}:} A multi-scale vision Transformer model employing a hierarchical structure for image processing.
    \item \textbf{Uniformer V2\cite{li2023uniformerv2}:} A Transformer architecture with a "Uniform Attention" mechanism to accelerate self-attention.
    \item \textbf{ST-GCN\cite{yan2018spatial}:} A network to represent human skeleton data as a graph, extracting spatial and temporal features.
    \item \textbf{2s-AGCN\cite{shi2019two}:} A two-stream architecture with separate GCN branches, building upon ST-GCN.
    \item \textbf{ST-GCN++\cite{duan2022pyskl}:} An improved version of ST-GCN, potentially addressing limitations or introducing enhancements.
    \item \textbf{PoseC3D\cite{duan2022revisiting}:} A 3D CNN architecture using skeleton data heat maps to recognize human activities.
\end{itemize}

\subsection{Comprehensive Performance comparision}
In this section, we compare our method with the above eight competing methods on our dataset MentalHAD. First, we use the above four metrics to compare the recognition performance on overall evaluation, the Setting I test, and the Setting II test. Then, we visualize confusion matrices of different methods with success and failure cases and give further discussion.

\subsubsection{State-of-the-Art Comparison}
We compare our proposed approach with the mentioned competing algorithms in terms of overall recognition performance. It's worth noting that these methods either incorporate scene information with human body data as RGB pixel input or discard scene information altogether, which cannot fully reveal the interaction relationship between humans and scene elements. Table \ref{comparison} summarizes the comprehensive performance of these algorithms on both all actions and abnormal actions. To ensure a fair comparison, we adopt the evaluation protocol outlined in section \ref{protocols}.

In Table \ref{comparison}, we find that our method has significant improvements over the other state-of-the-art methods on MentalHAD. For experiments on all actions, including both abnormal and normal actions, the results of the overall evaluation, Setting I test, and Setting II test show that compared with the best competing method (PoseC3D), our method achieves 11.8\%, 6.4\%, 14.4\% gains at $F_m$ on overall performance, subject- and scene-generalization ability, respectively, all reaching a high level of accuracy over 92\%. Especially for the key abnormal actions, our targeted optimization makes our method achieve a great leap of about 30.8\%, 38.5\% at $F_m$ on the overall performance and scene generalization ability of such actions and well maintaining the good subject generalization ability brought by skeleton features, exceeding the $F_m$ of the best competitive methods 15.1\%. Furthermore, the greater improvements in abnormal actions compared to all actions also confirm the challenges we mentioned in the introduction truly lie in abnormal actions. The great performance in the Setting II test also reveals the application potential in un-seen scenes, including the ward-like scene III in the Testing Subset 2.

In addition, the skeleton-based methods generally perform better than the RGB-based methods in all three tests, especially in the Setting I and Setting II tests. For the best method of the two types, the former is 10.3\%, 20.4\% higher than the latter's $F_m$ index in the Setting I and II tests. To a certain extent, this reveals the advantages of the skeleton-based method in generalization ability and confirms the characteristics of high robustness of skeleton modal data, thus preliminarily proving the superiority of our choice of baseline using skeleton data.

\subsubsection{Qualitative Evaluation}
To better demonstrate the excellent performance of our proposed method, we show the qualitative predicted results in Fig. \ref{conf}, where we visualize the confusion matrices of our proposed method and two competing methods on the overall testing set of MentalHAD. 

From Fig. \ref{cuniformer} and Fig. \ref{cp3d}, we notice significant confusion in predictions from the two competing methods, especially for abnormal actions. As an RGB-based method, Uniformer V2 treats human and scene pixels equally, blurring the interaction between them, while PoseC3D, as a skeleton-based method, disregards scene information entirely, leading to confusion mainly for abnormal actions. Additionally, RUniformer V2, when trained with limited scene diversity, tend to over-rely on scene information, causing confusion between normal actions across subjects and scenes in Fig. \ref{cuniformer}.

In contrast, Fig. \ref{cours} demonstrates minimal confusion in our method compared to Fig. \ref{cuniformer} and Fig. \ref{cp3d}. Although some confusion exists in action pairs like "climbing and hitting", and "hand wave and run", risky abnormal actions like climbing walls and hitting windows are accurately distinguished, avoiding misreporting and underreporting in practical application.

\subsubsection{Failure Cases Analysis}
\begin{figure}[t]
\centering
\subfloat[Setting I]{
\includegraphics[width=0.643\linewidth]{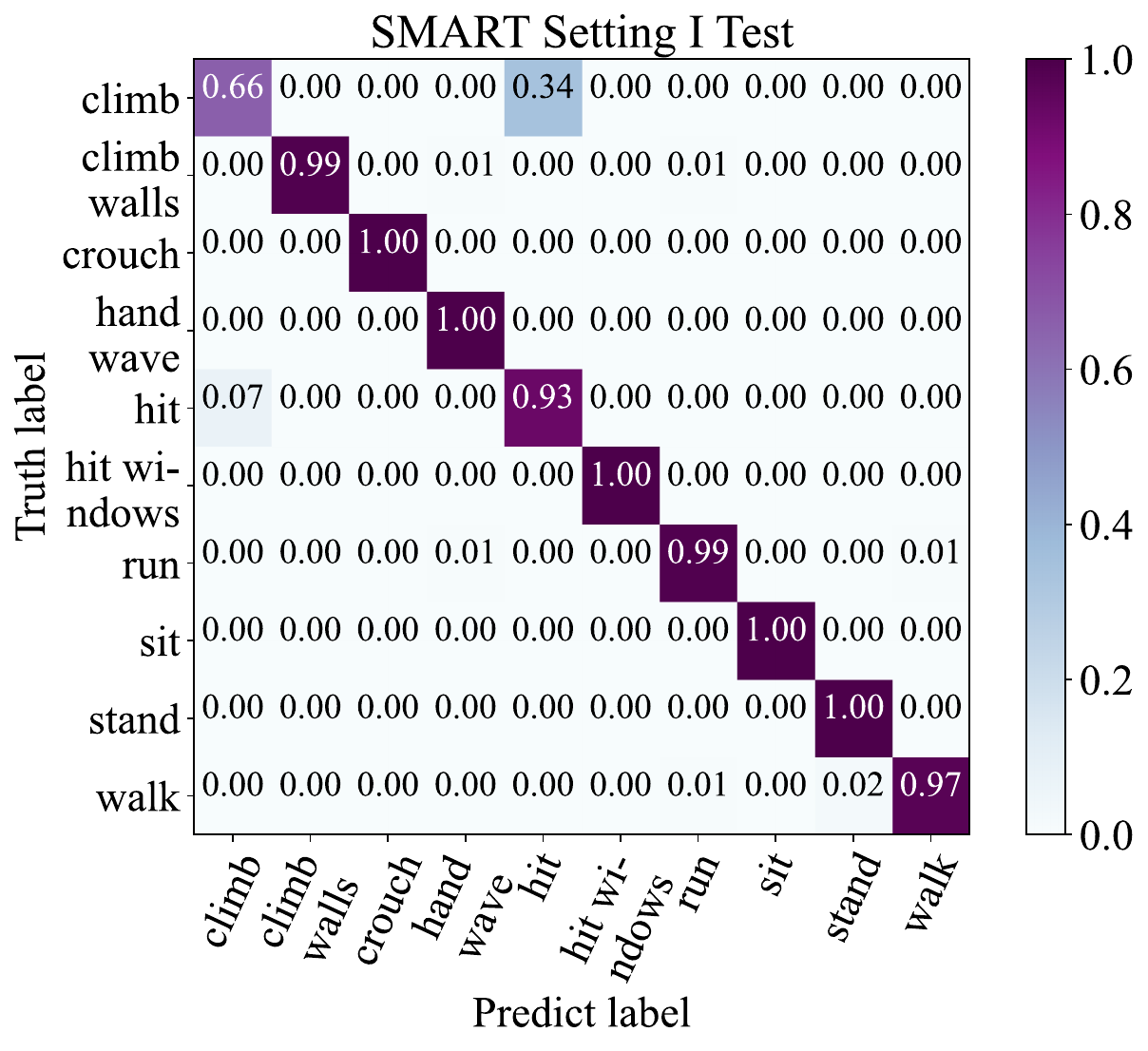}
\label{courssubject}}\\
\subfloat[Setting II]{
\includegraphics[width=0.643\linewidth]{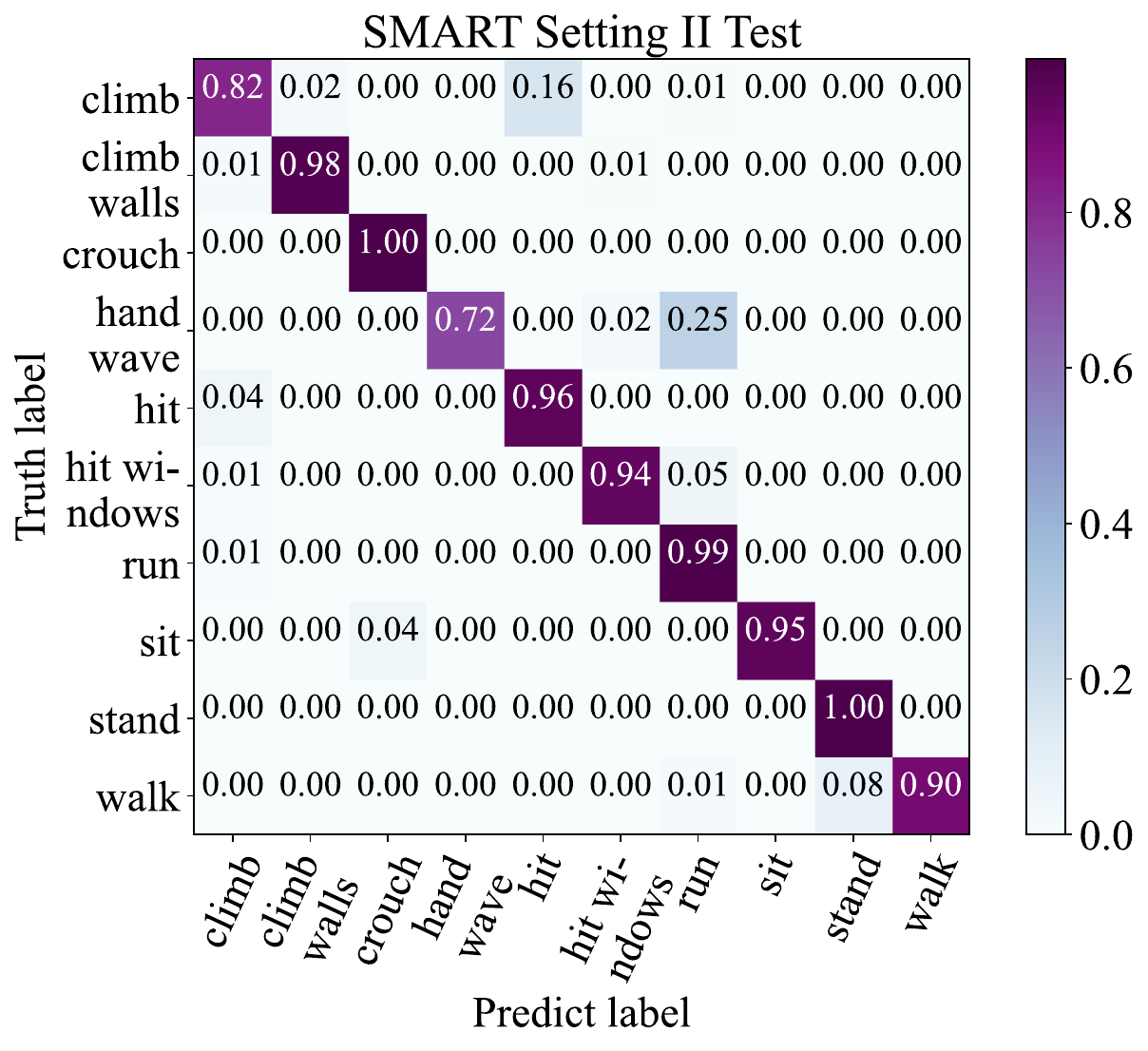}\label{coursscene}}
\caption{The confusion matrix of our method under Setting I and Setting II. (a) Setting I. (b) Setting II.}
\label{courssetting}
\end{figure}

To prove the reliability of our method, we further analyze the failure cases encountered in the above tests. Fig. \ref{courssetting} further shows the confusion matrices of SMART in the Setting I and Setting II tests.

In Fig. \ref{courssubject}, confusion primarily lies in the "climbing and hitting" action pair. This may arise from similar poses in these actions caused by low action quality from certain subjects, affecting the quality of skeleton features and subsequently fused features. Additionally, introducing scene interaction features could introduce noise, impacting the raw robustness of skeleton features and decreasing subject generalizability to some extent. Fig. \ref{coursscene} and Fig. \ref{cours} show similar confusion in the "climbing and hitting", "hand wave and run" action pairs from sequences in both Setting I and Setting II tests. Confusion between "climbing" and "hitting" may have similar reasons in Setting I. Confusion between "hand wave" and "run" may arise from "hand wave" data with large motion ranges on small screens, causing human statistics center movement patterns to resemble slow "run" actions. Collecting motion data with various human body proportions in images will address it.

\subsection{Ablation Study}
In this section, we quantitatively verify the effectiveness of our different modules and methods and give qualitative representations to assist analysis. We conduct the ablation study based on the baseline, module composition, scene information selection, scene feature extraction method, and fusion method. 

\subsubsection{Impact of Baseline}
\begin{figure}[t]
\centering
\includegraphics[width=1\linewidth]{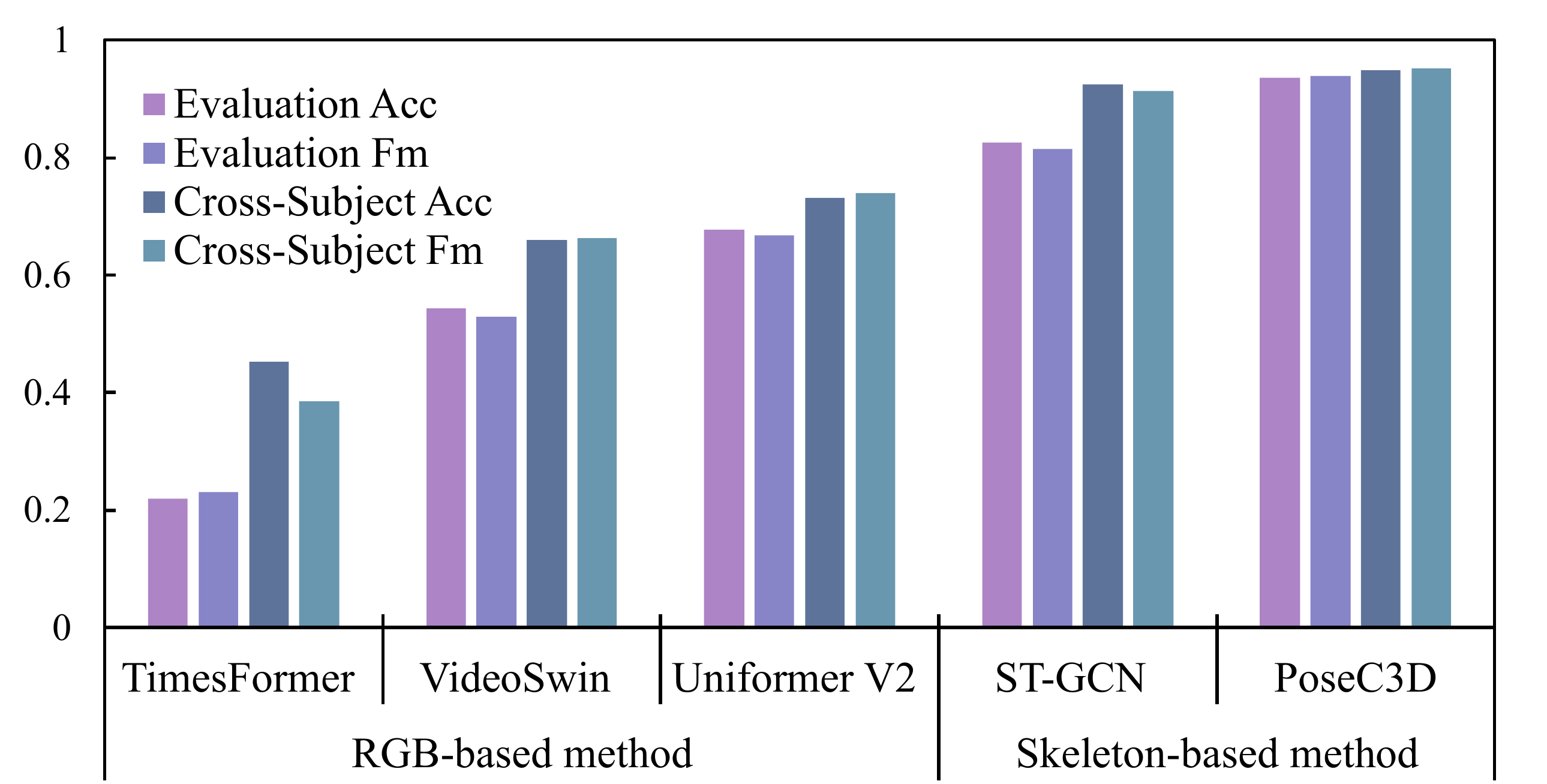}
\caption{The overall performance and subject generalization ability with different baselines}
\label{baselinef}
\end{figure}

To demonstrate the superiority of the baseline we selected, we substitute the skeleton feature extractor of PoseC3D with feature extractors from other methods. Fig. \ref{baselinef} visually presents the impact of different baselines on overall performance and subject generalizability.

Fig. \ref{baselinef} highlights that skeleton-based methods (ST-GCN \& PoseC3D) outperform RGB-based methods significantly in the Setting I test. This underscores the robustness of skeleton features across subjects and emphasizes the contribution of selecting skeleton-based methods to enhance subject generalization ability. Furthermore, comparing the two skeleton-based methods, PoseC3D exhibits stronger subject generalization ability and overall performance improvement than ST-GCN. This suggests that the quality of skeleton features profoundly influences the final fused feature quality. These findings affirm the superiority of selecting PoseC3D as our baseline.

\subsubsection{Impact of Module Composition}
\label{comp}
\begin{figure}[t]
\centering
\subfloat[No Scene Information]{
\includegraphics[width=0.643\linewidth]{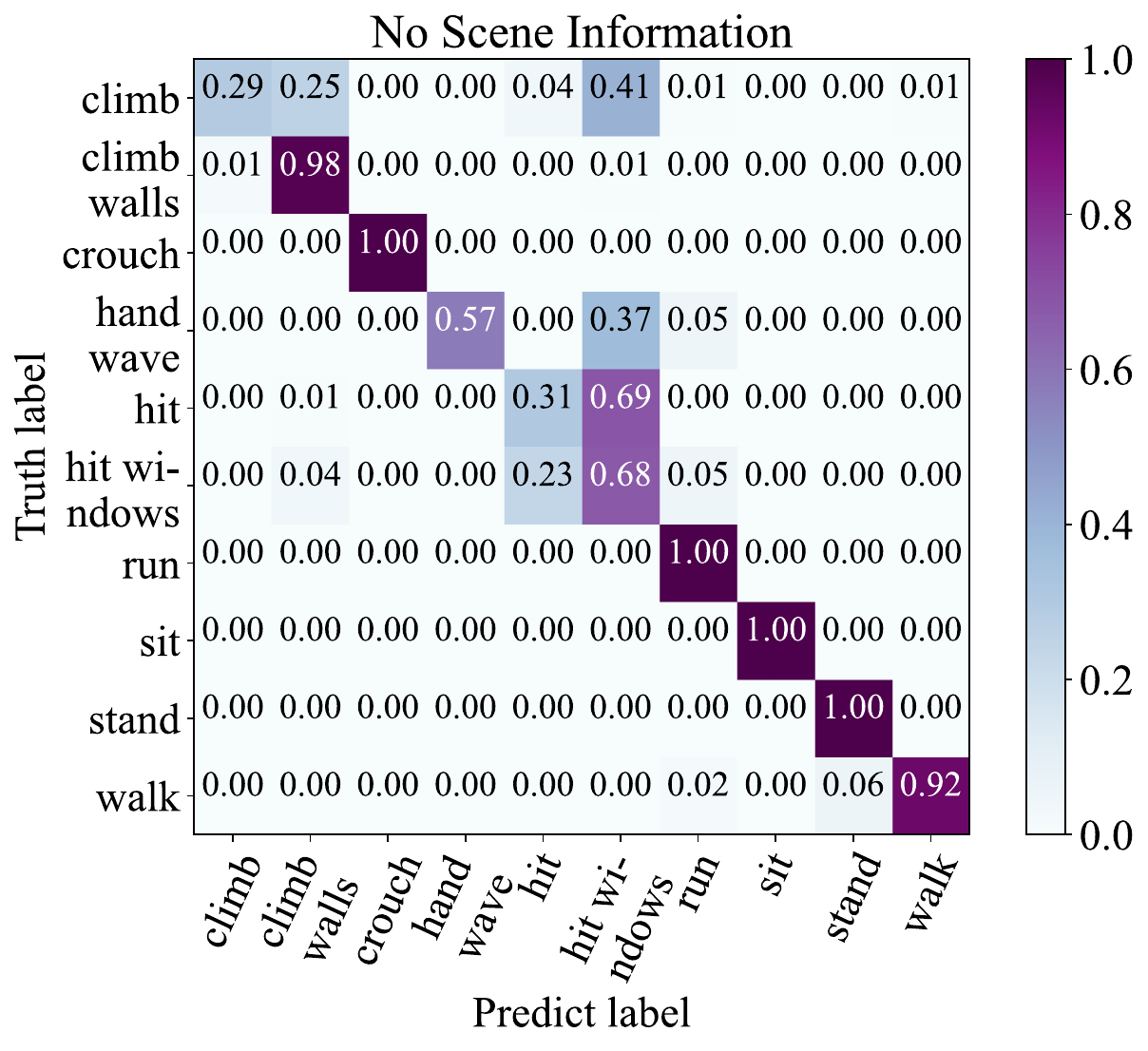}}\\
\subfloat[No Fusion Design]{
\includegraphics[width=0.643\linewidth]{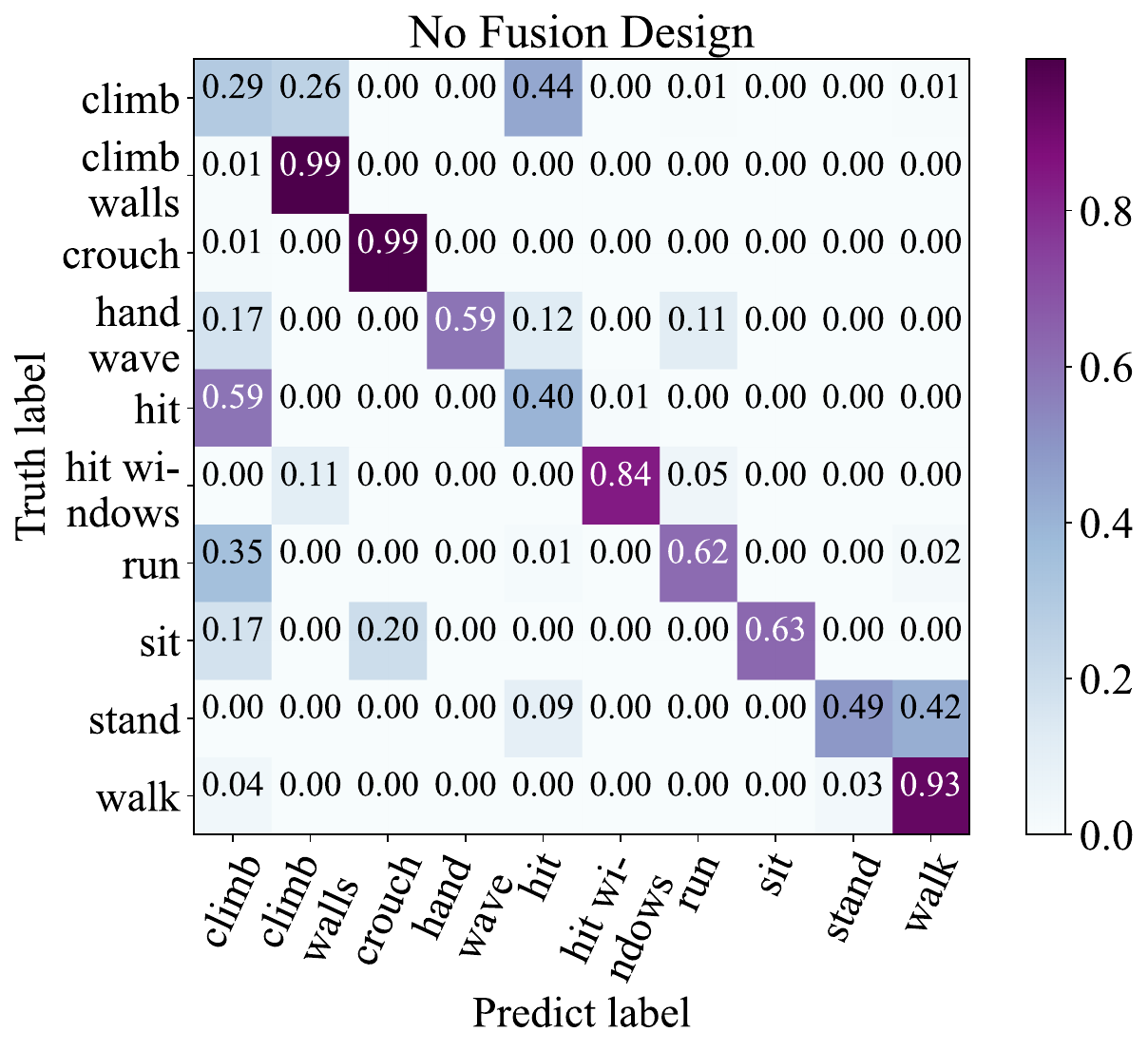}}
\caption{The confusion matrices of two frameworks with different module compositions on MentalHAD in Setting II test. (a) No Scene Perception Module. (b) No fusion method design in the Fusion and Classification Module.}
\label{cmmodule}
\end{figure}

To evaluate the effectiveness of each module, we conduct an ablation study on different module compositions. We remove the Scene Perception Module and the fusion design from the Fusion and Classification Module. The former results in the network solely relying on skeleton features for classification, while removing the fusion design leads us to use a simple concatenation method for feature fusion. Since the Scene Perception Module influences scene generalization ability, we focus on the overall evaluation and the Setting II test. Table \ref{composition} summarizes metrics with each module composition, and Fig. \ref{cmmodule} displays confusion matrices of the Setting II test.
\begin{table}[t]
    \centering
    \caption{Ablation study with different module composition}
    \label{composition}
    \begin{threeparttable}
    \resizebox{\linewidth}{!}{
    \begin{tabular}{c|>{\centering\arraybackslash}p{1.3cm}>{\centering\arraybackslash}p{1.3cm}|>{\centering\arraybackslash}p{1.3cm}>{\centering\arraybackslash}p{1.3cm}}
      \toprule
      \multirow{2}*{\textbf{\raisebox{-6pt}{\makecell[c]{Info}}}} & \multicolumn{2}{c}{\textbf{Overall}} \vline & \multicolumn{2}{c}{\textbf{Setting II}}\\ \cmidrule(l{1pt}r{0pt}){2-5}
        &\textbf{$Acc$} & \textbf{$F_m$} &\textbf{$Acc$} & \textbf{$F_m$}\\
        \midrule
      No Scene & 82.8\% &81.8\% & 77.5\%& 77.3\%\\
      No Fusion & 74.5\%& 75.5\%&67.8\%&69.0\%\\
      Ours & \textbf{93.6\%} & \textbf{93.9\%}& \textbf{93.2\%}&\textbf{92.7\%}\\
      \bottomrule
    \end{tabular}}
  \begin{tablenotes}
  \footnotesize
  \item[*] No scene denotes the module composition without the Scene Perception Module; No Fusion denotes the module composition without fusion design in the Fusion and Classification Module.
  \end{tablenotes}
  \end{threeparttable}
\end{table}

Table \ref{composition} underscores the effectiveness of our module composition in overall and Setting II tests. The removal of the Scene Perception Module leads to a significant decrease in two $F_m$ scores compared to our method, indicating the essential role of scene information in final features. Additionally, employing concatenation as the fusion method introduces more noise, resulting in poorer performance due to the absence of effective feature fusion methods. Qualitatively, Fig. \ref{cmmodule} illustrates that the absence of scene information primarily leads to confusion among abnormal actions, while the simple fusion method causes confusion for all actions. This underscores the importance of scene information for abnormal action recognition and highlights how an unsuitable fusion method introduces redundant noise from scene information, confusing the skeleton feature of all actions. These experiments prove the necessity and significance of the Scene Perception Module and an appropriate fusion method, inspiring further detailed ablation studies within each module.

\subsubsection{Impact of Scene Information Selection}
\begin{figure}[t]
\centering
\subfloat[None]{\includegraphics[width=0.48\linewidth]{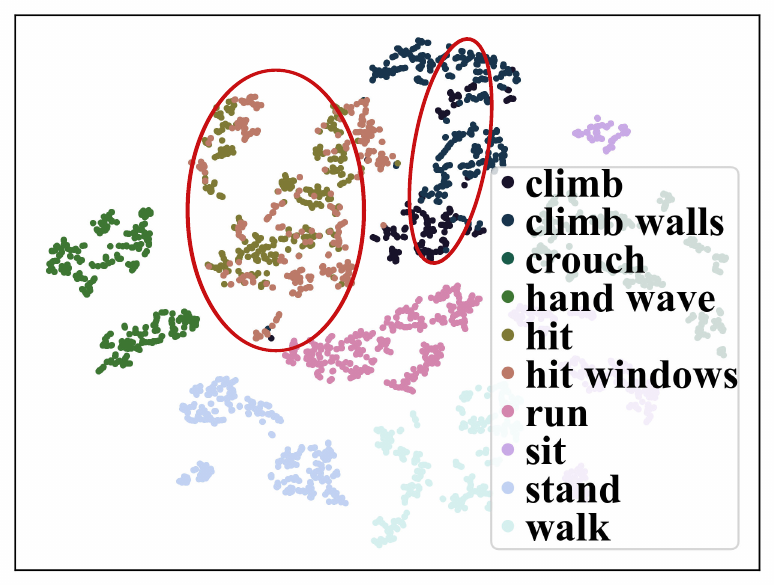}}
\hfil
\subfloat[Only Depth]{
\includegraphics[width=0.48\linewidth]{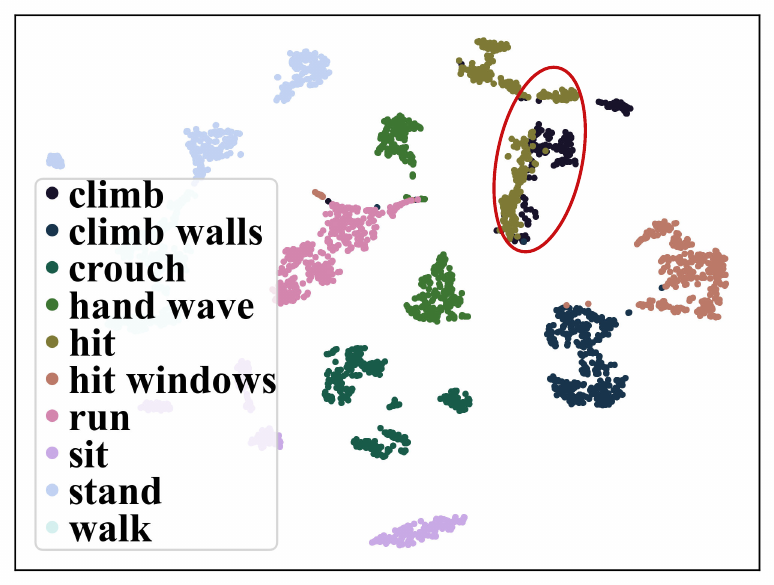}}\\
\subfloat[Only Mask]{
\includegraphics[width=0.48\linewidth]{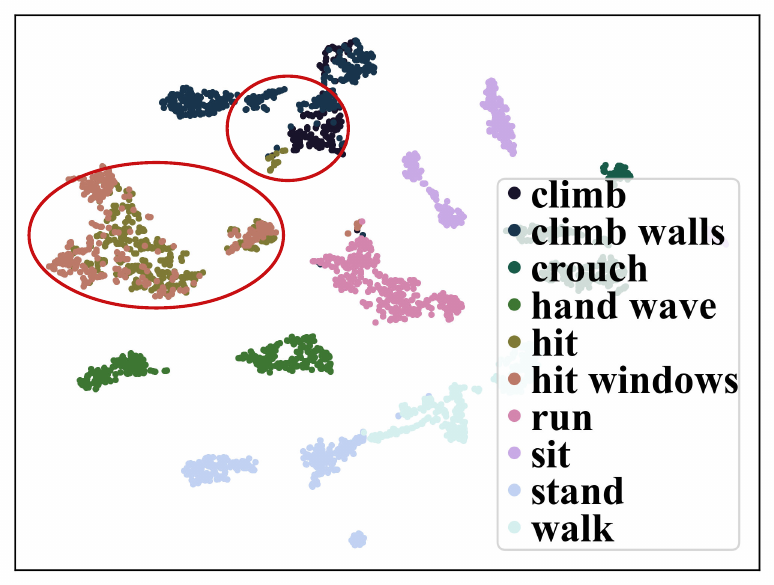}}
\hfil
\subfloat[Ours]{
\includegraphics[width=0.48\linewidth]{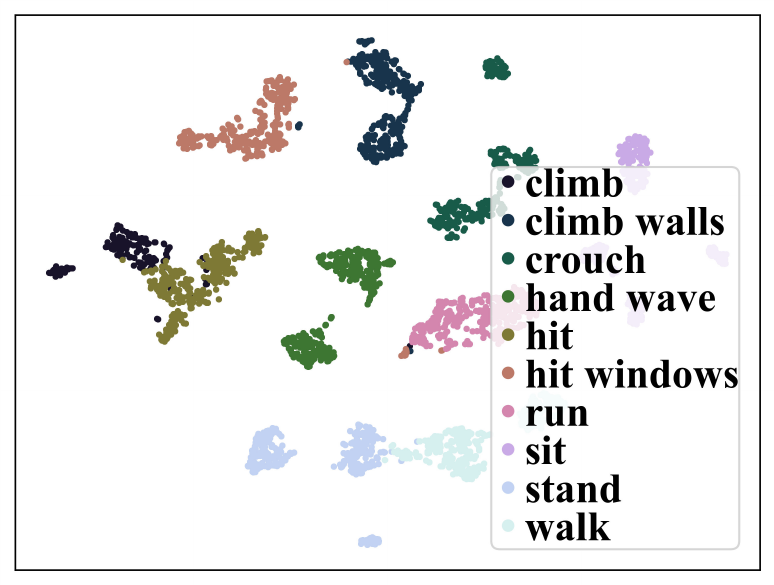}}
\caption{The t-SNE visualization of SMART with different scene information selection on MentalHAD Subset 2. (a) With no scene information. (b) Only with the depth as scene information. (c) Only with the mask as scene information. (d) With both depth and mask as scene information.}
\label{tsneselectionscene}
\end{figure}

Considering the importance of the Scene Perception Module, we extract suitable scene information for subsequent feature extraction. To validate the selection of our semantic depth and mask data, we compare four cases: depth data only, mask data only, both depth and mask data, and neither. Table \ref{selection} shows the metrics for each selection, and Fig. \ref{tsneselectionscene} compares t-SNE plots of the final features in the corresponding Setting II test.

\begin{table}[t]
    \centering
    \caption{Ablation study with different scene information}
    \label{selection}
    \begin{threeparttable}
    \resizebox{\linewidth}{!}{
    \begin{tabular}{c|>{\centering\arraybackslash}p{1.3cm}>{\centering\arraybackslash}p{1.3cm}|>{\centering\arraybackslash}p{1.3cm}>{\centering\arraybackslash}p{1.3cm}}
      \toprule
      \multirow{2}*{\textbf{\raisebox{-6pt}{\makecell[c]{Info}}}}
        &\multicolumn{2}{c}{\textbf{Overall}}\vline& \multicolumn{2}{c}{\textbf{Setting II}}\\ \cmidrule(l{1pt}r{0pt}){2-5}
        &\textbf{$Acc$} & \textbf{$F_m$} &\textbf{$Acc$} & \textbf{$F_m$}\\
        \midrule
      None(1) & 82.8\% &81.8\%& 77.5\%& 77.3\%\\
      D(2) & 88.4\% & 87.5\%& 84.7\%&82.4\%\\
      M(3) & 83.0\% & 83.0\%& 75.2\%&74.9\%\\
      Ours(4) & \textbf{93.6\% }& \textbf{93.1\%}& \textbf{93.1\%}&\textbf{92.6\%}\\
      \bottomrule
    \end{tabular}}
  \begin{tablenotes}
  \footnotesize
  \item[*] Info denotes the Input scene information selection options, None denotes no scene information, D denotes only depth data as scene information, M denotes only mask data is introduced, and Ours denotes depth and mask data as scene information.
  \end{tablenotes}
  \end{threeparttable}
\end{table}

From Table \ref{selection}, our scene information selection significantly improves performance. Comparing the performance of (2), (3), and (1), our method outperforms others in both overall evaluation and the Setting II test, as indicated by higher $Acc$ and $F_m$ scores. Depth and mask data contribute to enhanced performance, with $F_m$ increasing by 5.7\% and 1.2\%, respectively. Depth information particularly improves scene generalization, increasing $F_m$ by 5.1\% in Setting II. Despite a slight decrease in metrics, mask information complements depth by aiding in semantic depth generation, ultimately improving performance. Compared to the baseline (1), our method achieves performance gains of 11.3\% and 15.3\% on $F_m$ for overall and scene generalization, respectively. Qualitative analysis from Fig. \ref{tsneselectionscene} further confirms the efficacy of our selection in enhancing action discrimination, particularly in scene generalization, reducing confusion between actions with similar scene information like climbing and hitting, and actions with similar skeletons like hitting and hitting windows.

\subsubsection{Impact of Human-Scene Interaction Feature Extraction Method}
The features obtained by different feature extraction methods differ in semantic completeness and noise for the same information. According to section \ref{spm}, SMART utilizes a self-attention network to extract the human motion trajectory feature and uses a Dual-Siamese Network to extract the human-scene interaction feature. To verify their effectiveness, we choose two networks with similar architecture to employ the ablation study, which is introduced as follows:

\begin{itemize}
    \item \textbf{SFE1:} Inspired by Two-stream Network\cite{yan2018spatial} and 3D-CNN\cite{duan2022revisiting}, we construct Two-stream 3D-CNN\cite{kondratyuk2021movinets} to extract depth and mask features.
    \item \textbf{SFE2:} Inspired by the self-attention and Siamese networks, we integrate them into this method. A self-attention network extracts motion trajectory features. A Siamese network captures interaction features from semantic masks. And a 3D-CNN extracts interaction features from semantic depth.
\end{itemize}

\begin{figure}[t]
\centering
\subfloat[SFE1]{
\includegraphics[width=0.643\linewidth]{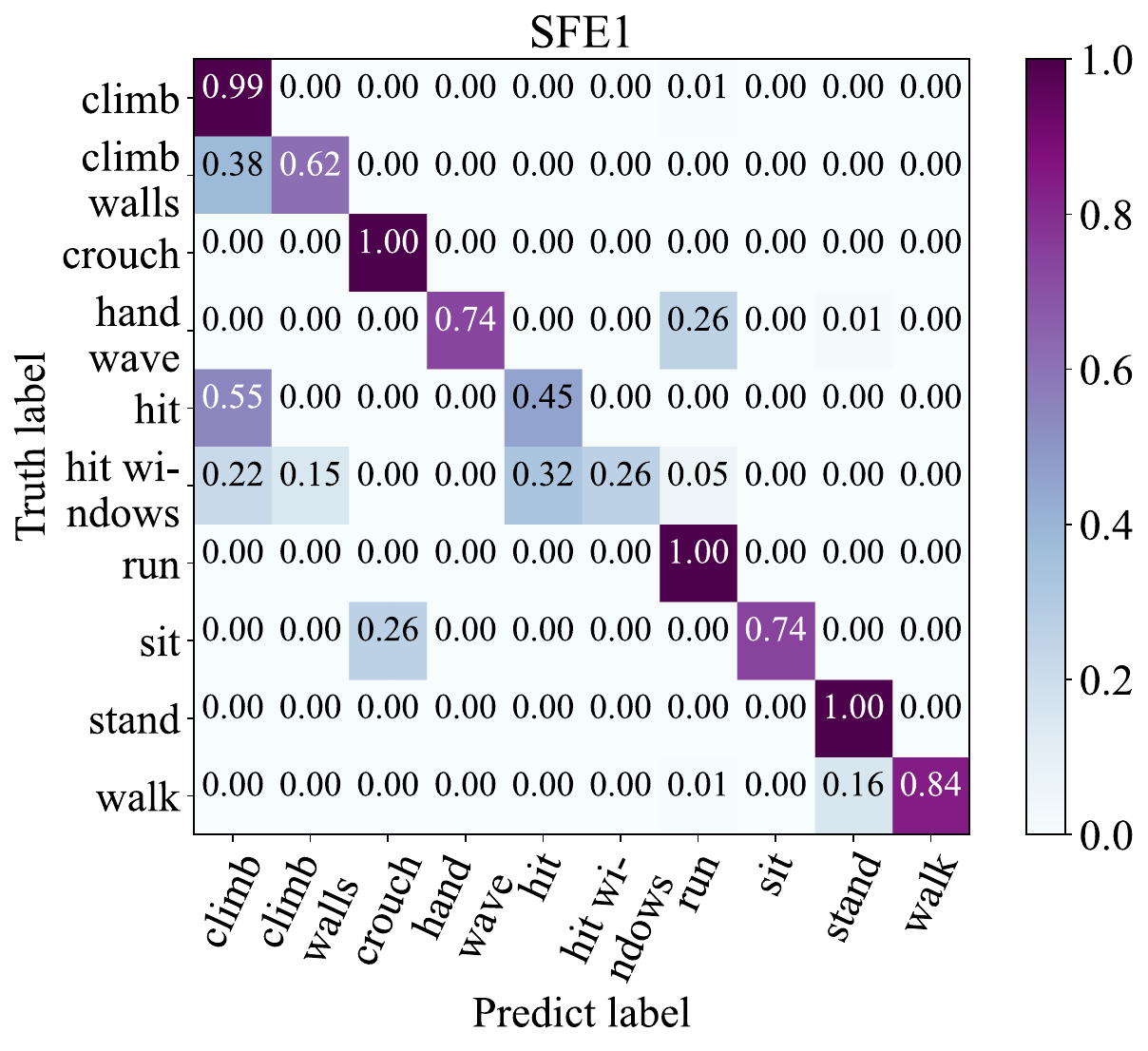}}\\
\subfloat[SFE2]{
\includegraphics[width=0.643\linewidth]{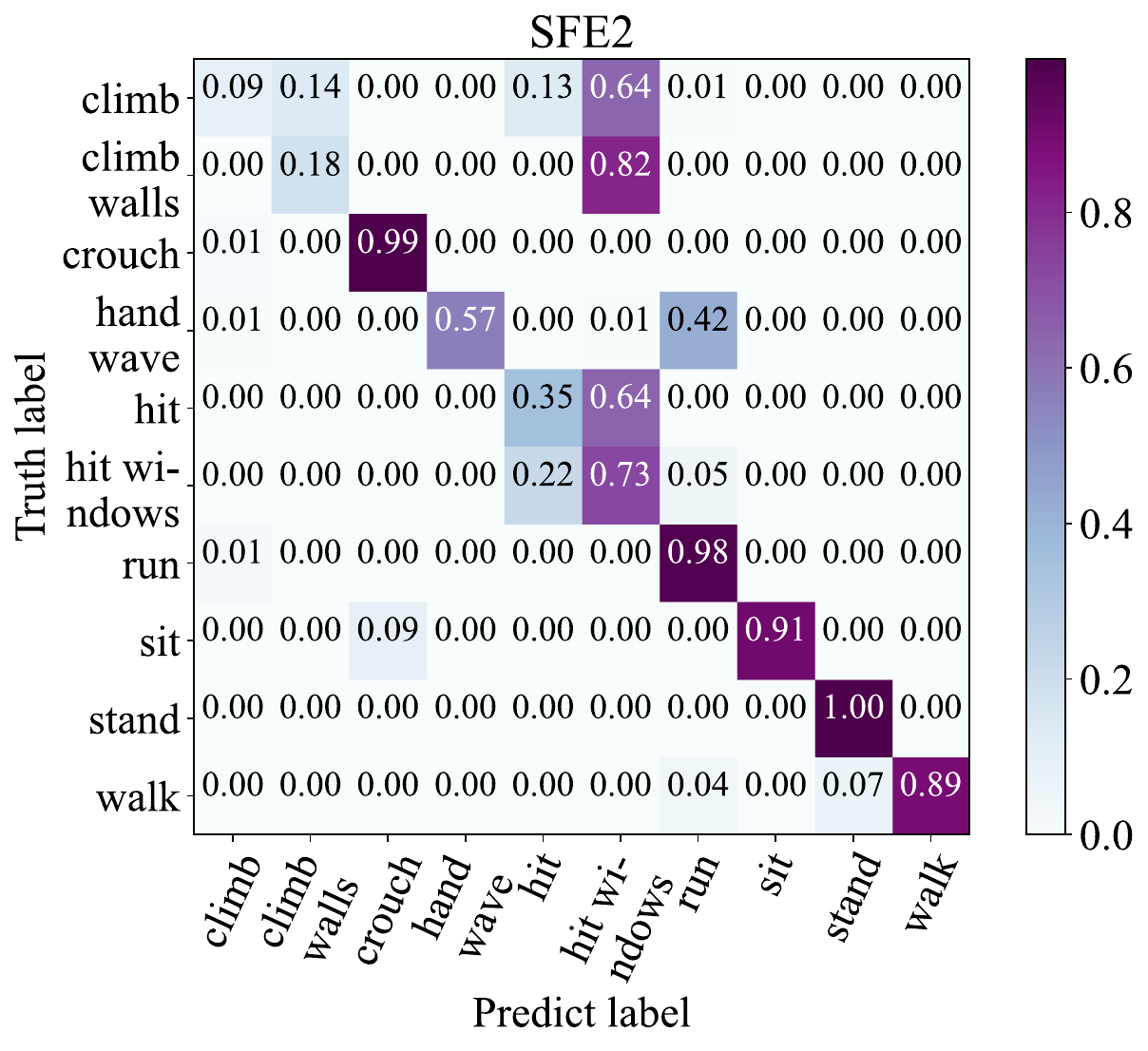}}
\caption{The confusion matrices of SMART with two human-scene interaction feature extraction methods in the Setting II test. (a) SFE1. (b) SFE2.}
\label{sfec}
\end{figure}

Table \ref{sfe} shows the metrics for each method. Figure \ref{sfec} compares the confusion matrices of different methods in the Setting II test. From Table \ref{sfe}, our method excels in both overall evaluation and the Setting II test. Among the three methods, the first one struggles due to limited data and lacks semantic guidance, hampering its performance. While the second method uses some semantic guidance, its network architecture restricts clear semantic features. Qualitatively, from Fig. \ref{sfec}, we observe that the first method's confusion spreads to more normal actions compared to the second method, which mainly confuses abnormal actions. This underscores the effectiveness of the second method's strategy of splitting features into two categories with clearer semantics. Our improved performance further validates this strategy and new network architecture.
\begin{table}[t]
  \begin{center}
    \caption{Ablation Study on Human-Scene Interaction Feature Extraction Method}
    \label{sfe}
    \begin{threeparttable}
    \resizebox{\linewidth}{!}{
    \begin{tabular}{c|>{\centering\arraybackslash}p{1.3cm}>{\centering\arraybackslash}p{1.3cm}|>{\centering\arraybackslash}p{1.3cm}>{\centering\arraybackslash}p{1.3cm}}
        \toprule
        \multirow{2}*{\textbf{\raisebox{-6pt}{\makecell[c]{Method}}}}
        &\multicolumn{2}{c}{\textbf{Overall}} \vline& \multicolumn{2}{c}{\textbf{Setting II}}\\ \cmidrule(l{1pt}r{0pt}){2-5}
        &\textbf{$Acc$} & \textbf{$F_m$}  &\textbf{$Acc$} & \textbf{$F_m$}\\
        \midrule
      SFE1&81.3\%& 81.0\%&76.4\%&75.4\%\\
      SFE2&77.9\%&77.9\%&66.9\%&65.5\%\\
      Ours&\textbf{93.6\%}& \textbf{93.1\%}&\textbf{93.2\%}&\textbf{92.7\%}\\
      \bottomrule
    \end{tabular}}
    \end{threeparttable}
  \end{center}
\end{table}

\subsubsection{Impact of Fusion Method}
The fusion method has been proved important in section \ref{comp}. In our method, we first obtain the human motion feature by fusing the human motion trajectory feature and the skeleton motion feature with channel attention. Then, the human-scene interaction feature is weighted with a scene weight network to decide the association degree between humans and scene elements for each action. Finally, the weighted human-scene interaction feature is fused with the human motion feature with another channel attention network. To prove the effectiveness of our fusion method, we conduct the ablation study on four fusion methods. The other three fusion methods are claimed as follows: 

\begin{itemize}
    \item \textbf{Concat:} Simply concatenating the human motion trajectory feature and human skeleton motion feature, as well as the human-scene interaction feature and the above fused feature.
    \item \textbf{Self-Attention\cite{10007033}:} Utilizing an attention mechanism to prioritize relevant parts of the simple-concatenated feature.
    \item \textbf{Channel-Attention\cite{wang2020eca}:} Applying attention across different channels within the concatenated feature vector to emphasize the most informative channels.
\end{itemize}

\begin{figure}[t]
\centering
\subfloat[Concat]{
\includegraphics[width=0.48\linewidth]{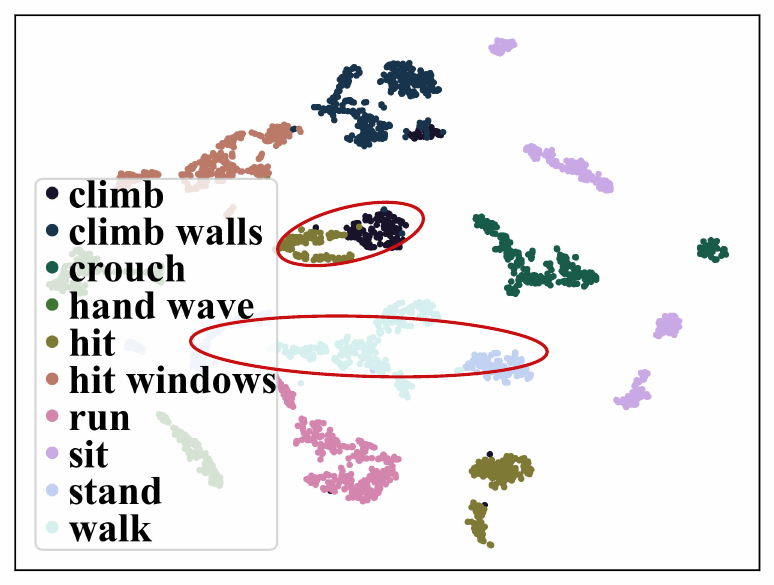}}
\hfil
\subfloat[Self-attention]{
\includegraphics[width=0.48\linewidth]{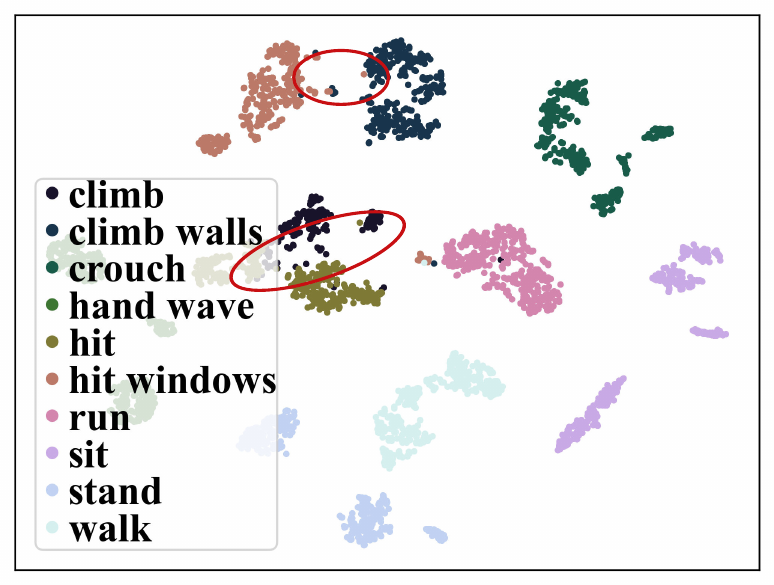}}\\
\subfloat[Channel-attention]{
\includegraphics[width=0.48\linewidth]{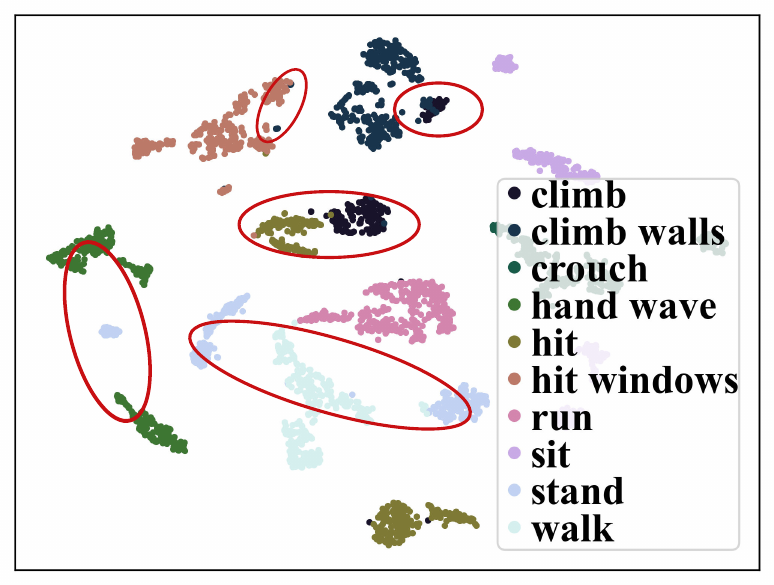}}
\hfil
\subfloat[Ours]{
\includegraphics[width=0.48\linewidth]{figure/SMART_test_cross_scene_tsne.pdf}}
\caption{The t-SNE of SMART with different fusion methods in the Setting II test. (a) Concat. (b) Self-Attention. (c) Channel-Attention. (d) Ours.}
\label{tsnefusion}
\end{figure}

Considering the module's influence on both subject and scene generalization due to its association with all features, we conduct the overall evaluation, the Setting I and II tests for comprehensive analysis. Table \ref{ffm} shows metrics for each method. Fig. \ref{tsnefusion} shows t-SNE plots of final action semantic features with different fusion methods in the Setting II test.

\begin{table}[t]
  \begin{center}
    \caption{Ablation Experiments on Feature Fusion Method}
    \label{ffm}
    \begin{threeparttable}
    \resizebox{\linewidth}{!}{
    \begin{tabular}{c|cc|cc|cc}
        \toprule
        \multirow{2}*{\textbf{\raisebox{-2pt}{\makecell[c]{Method}}}}
        &\multicolumn{2}{c}{\textbf{Overall}} \vline & \multicolumn{2}{c}{\textbf{Setting I}} \vline& \multicolumn{2}{c}{\textbf{Setting II}}\\ \cmidrule(l{1pt}r{0pt}){2-7}
        &\textbf{$Acc$} & \textbf{$F_m$} & \textbf{$Acc$} & \textbf{$F_m$} &\textbf{$Acc$} & \textbf{$F_m$}\\
        \midrule
      Concat & 74.5\%& 75.5\%& 92.9\%& 92.3\%&67.8\%&69.0\%\\
      SA\cite{10007033}& 85.8\%& 85.4\%& 90.1\%& 90.0\%&85.6\%&85.2\%\\
      CA\cite{wang2020eca}& 76.4\%& 77.0\%& 92.1\%& 91.1\%&70.2\%&70.8\%\\
      \textbf{Ours}& \textbf{93.6\%} & \textbf{93.1\%}& \textbf{94.9\%} & \textbf{95.1\%}& \textbf{93.1\%}&\textbf{92.6\%}\\
      \bottomrule
    \end{tabular}}
    \begin{tablenotes}
        \footnotesize
        \item[*] SA denotes Self-Attention, CA denotes Channel-Attention.
      \end{tablenotes}
    \end{threeparttable}
  \end{center}
\end{table}

From Table \ref{ffm}, our fusion method positively affects performance improvement. With the results of $Acc$ and $F_m$, our method performs best on the Setting I and II tests, where $F_m$ has increased by 2.8\% and 23.6\%, respectively, compared to the method Concat. Both the subject and scene generalization abilities are improved with our fusion method. Besides, from Fig. \ref{tsnefusion}, we could find the effectiveness of the scene weight network. Compared with the Channel-attention method, our method has less confusion among normal actions. This indicates our scene weight network determines the weights of introduced human-scene interaction features for each action. This helps SMART avoid using meaningless scene clues to identify actions unrelated to the scene, thus improving the scene generalization ability of all actions.

\subsection{Practical Deployment}
To verify our framework's practical effectiveness for hospital multi-way monitoring on a single GPU, we deploy a simplified version in a mental health center. This version employs a monocular camera positioning method for scene perception, enabling a preliminary test of our pipeline. While clinical data is confidential due to privacy, property, and ethical concerns, the center's positive feedback indicates the framework's ability to identify human behavior in real-time from surveillance video streams on hospital servers, to some extent, affirming the feasibility of our framework in applications.

\section{Conclusion and Future Works}
This paper proposes a novel scene-motion-aware human action recognition framework for both normal and abnormal actions primarily occurring in the mental disorder group, significantly improving action recognition performance compared to state-of-the-art methods. Facing data scarcity on abnormal actions, we construct a HAR dataset that includes six normal actions and four abnormal actions. Based on the motion perception of the skeleton-based HAR, our method uniquely introduces a scene perception module to extract human-scene interaction and human motion trajectory features, which enhance the representation of action semantics. A multi-stage fusion network is further proposed for optimal feature utilization. The effectiveness of each proposed method is verified with abundant qualitative ablation studies and quantitative analysis. Moreover, compared with the best state-of-the-art algorithm, our method improves the recognition performance ($F_m$) by 11.8\%, 6.4\%, 14.4\% in the overall evaluation, Setting I test and Setting II test on all actions and 30.8\%, 15.1\%, 38.5\% on abnormal actions. In general, our method has been proven to have high subject and scene generalizability for recognizing actions of the mental disorder group, making it possible to migrate our algorithm to the mental disorder group in medical settings for IoT-based smart healthcare. 

For future work, we plan to deploy our framework in the mental health center to comprehensively evaluate and optimize its practical application performance. We will also integrate the trainable scene information extractor into our framework to construct an end-to-end framework. Additionally, fine-tuning the model with abnormal action data from patients to improve adaptability in real scenes. Lastly, novel methodologies such as transformers and state space models will be explored to enhance subject and scene generalizability further.

\bibliographystyle{IEEEtran}
\bibliography{ref}

\begin{thebibliography}{10}
\providecommand{\url}[1]{#1}
\csname url@samestyle\endcsname
\providecommand{\newblock}{\relax}
\providecommand{\bibinfo}[2]{#2}
\providecommand{\BIBentrySTDinterwordspacing}{\spaceskip=0pt\relax}
\providecommand{\BIBentryALTinterwordstretchfactor}{4}
\providecommand{\BIBentryALTinterwordspacing}{\spaceskip=\fontdimen2\font plus
\BIBentryALTinterwordstretchfactor\fontdimen3\font minus \fontdimen4\font\relax}
\providecommand{\BIBforeignlanguage}[2]{{%
\expandafter\ifx\csname l@#1\endcsname\relax
\typeout{** WARNING: IEEEtran.bst: No hyphenation pattern has been}%
\typeout{** loaded for the language `#1'. Using the pattern for}%
\typeout{** the default language instead.}%
\else
\language=\csname l@#1\endcsname
\fi
#2}}
\providecommand{\BIBdecl}{\relax}
\BIBdecl

\bibitem{hao2021end}
Y.~Hao \emph{et~al.}, ``An end-to-end human abnormal behavior recognition framework for crowds with mentally disordered individuals,'' \emph{IEEE J. Biomed. Health. Inf.}, vol.~26, no.~8, pp. 3618--3625, 2021.

\bibitem{pei2021mars}
L.~Pei \emph{et~al.}, ``Mars: Mixed virtual and real wearable sensors for human activity recognition with multidomain deep learning model,'' \emph{IEEE Internet Things J.}, vol.~8, no.~11, pp. 9383--9396, 2021.

\bibitem{hu2024end}
D.~Hu, Y.~Fang, J.~Cao, T.~Jiang, and F.~Gao, ``An end-to-end vision-based seizure detection with a guided spatial attention module for patient detection,'' \emph{IEEE Internet Things J.}, 2024.

\bibitem{tan2023quantitative}
Y.~Tan, W.~Liu, H.-J. Sun, and P.~Wang, ``Quantitative measurement of parkinson’s disease gait based on the rehabilitation monitoring robot,'' \emph{IEEE Trans. Instrum. Meas.}, 2023.

\bibitem{adil2024healthcare}
M.~Adil \emph{et~al.}, ``Healthcare internet of things: Security threats, challenges and future research directions,'' \emph{IEEE Internet Things J.}, 2024.

\bibitem{duan2022revisiting}
H.~Duan, Y.~Zhao, K.~Chen, D.~Lin, and B.~Dai, ``Revisiting skeleton-based action recognition,'' in \emph{Proc. IEEE Conf. Comput. Vis. Pattern Recog.}, 2022, pp. 2969--2978.

\bibitem{ahmed2021deep}
I.~Ahmed, G.~Jeon, and F.~Piccialli, ``A deep-learning-based smart healthcare system for patient’s discomfort detection at the edge of internet of things,'' \emph{IEEE Internet Things J.}, vol.~8, no.~13, pp. 10\,318--10\,326, 2021.

\bibitem{xia2024timestamp}
S.~Xia \emph{et~al.}, ``Timestamp-supervised wearable-based activity segmentation and recognition with contrastive learning and order-preserving optimal transport,'' \emph{IEEE Trans. Mob. Comput.}, 2024.

\bibitem{yang2024mmbat}
J.~Yang, S.~Xia, Y.~Song, Q.~Wu, and L.~Pei, ``mmbat: A multi-task framework for mmwave-based human body reconstruction and translation prediction,'' in \emph{ICASSP IEEE Int Conf Acoust Speech Signal Process Proc}.\hskip 1em plus 0.5em minus 0.4em\relax IEEE, 2024, pp. 8446--8450.

\bibitem{waqar2023simulation}
S.~Waqar and M.~P{\"a}tzold, ``A simulation-based framework for the design of human activity recognition systems using radar sensors,'' \emph{IEEE Internet Things J.}, 2023.

\bibitem{wang2023videomae}
L.~Wang \emph{et~al.}, ``Videomae v2: Scaling video masked autoencoders with dual masking,'' in \emph{Proc. IEEE Conf. Comput. Vis. Pattern Recog.}, 2023, pp. 14\,549--14\,560.

\bibitem{chang2022contrastive}
S.~Chang, Y.~Li, S.~Shen, J.~Feng, and Z.~Zhou, ``Contrastive attention for video anomaly detection,'' \emph{IEEE Trans. Multimedia}, vol.~24, p. 4067, 2022.

\bibitem{chen2023fall}
D.~Chen, A.~B. Wong, and K.~Wu, ``Fall detection based on fusion of passive and active acoustic sensing,'' \emph{IEEE Internet Things J.}, 2023.

\bibitem{wu2023video}
L.~Wu \emph{et~al.}, ``Video-based fall detection using human pose and constrained generative adversarial network,'' \emph{IEEE Trans. Circuits Syst. Video Technol.}, 2023.

\bibitem{zhang2023deep}
Y.~Zhang \emph{et~al.}, ``Deep learning based abnormal behavior detection for elderly healthcare using consumer network cameras,'' \emph{IEEE Trans. Consum. Electron.}, 2023.

\bibitem{chen2019semisupervised}
K.~Chen \emph{et~al.}, ``A semisupervised recurrent convolutional attention model for human activity recognition,'' \emph{IEEE Trans. Neural Networks Learn. Syst.}, vol.~31, no.~5, pp. 1747--1756, 2019.

\bibitem{bulling2014tutorial}
A.~Bulling, U.~Blanke, and B.~Schiele, ``A tutorial on human activity recognition using body-worn inertial sensors,'' \emph{ACM Comput Surv}, vol.~46, no.~3, pp. 1--33, 2014.

\bibitem{zhang2023dynamic}
Y.~Zhang \emph{et~al.}, ``Dynamic inertial poser (dynaip): Part-based motion dynamics learning for enhanced human pose estimation with sparse inertial sensors,'' \emph{arXiv preprint arXiv:2312.02196}, 2023.

\bibitem{sun2022human}
Z.~Sun \emph{et~al.}, ``Human action recognition from various data modalities: A review,'' \emph{IEEE Trans. Pattern Anal. Mach. Intell.}, vol.~45, no.~3, pp. 3200--3225, 2022.

\bibitem{zeng2022inversionnet3d}
Q.~Zeng, S.~Feng, B.~Wohlberg, and Y.~Lin, ``Inversionnet3d: Efficient and scalable learning for 3-d full-waveform inversion,'' \emph{IEEE Trans. Geosci. Remote Sens.}, vol.~60, p. 3135354, 2022.

\bibitem{fiehler2023spatial}
K.~Fiehler and H.~Karimpur, ``Spatial coding for action across spatial scales,'' \emph{Nat. Rev. Psychol.}, vol.~2, no.~2, pp. 72--84, 2023.

\bibitem{10230011}
J.~Li, Z.~Zhao, J.~Yang, H.~Chu, and Q.~Li, ``Self-constructing temporal excitation graph for skeleton-based action recognition,'' \emph{IEEE Sens. J.}, vol.~23, no.~19, pp. 23\,079--23\,091, 2023.

\bibitem{10413307}
J.~Lu \emph{et~al.}, ``Dual-excitation spatial–temporal graph convolution network for skeleton-based action recognition,'' \emph{IEEE Sens. J.}, vol.~24, no.~6, pp. 8184--8196, 2024.

\bibitem{qiu2021skeleton}
J.~Qiu, X.~Yan, W.~Wang, W.~Wei, and K.~Fang, ``Skeleton-based abnormal behavior detection using secure partitioned convolutional neural network model,'' \emph{IEEE J. Biomed. Health. Inf.}, vol.~26, no.~12, pp. 5829--5840, 2021.

\bibitem{arifoglu2019detection}
D.~Arifoglu and A.~Bouchachia, ``Detection of abnormal behaviour for dementia sufferers using convolutional neural networks,'' \emph{Artif. Intell. Med.}, vol.~94, pp. 88--95, 2019.

\bibitem{sultani2018real}
W.~Sultani, C.~Chen, and M.~Shah, ``Real-world anomaly detection in surveillance videos,'' in \emph{Proc. IEEE Conf. Comput. Vis. Pattern Recog.}, 2018, pp. 6479--6488.

\bibitem{chapron2019highly}
K.~Chapron, P.~Lapointe, K.~Bouchard, and S.~Gaboury, ``Highly accurate bathroom activity recognition using infrared proximity sensors,'' \emph{IEEE J. Biomed. Health. Inf.}, vol.~24, no.~8, pp. 2368--2377, 2019.

\bibitem{9853267}
J.~Hao, M.~Gong, L.~Wang, H.~Xiong, and X.~Huang, ``Edge computing based abnormal behavior learning for mental disorder detection through uav surveillance,'' in \emph{Proc. - Int. Symp. Electr., Electron. Inf. Eng., ISEEIE}, 2022, pp. 25--29.

\bibitem{6359337}
X.-g. Chen, J.~Liu, and H.~Liu, ``Will scene information help realistic action recognition?'' in \emph{Proc. World Congr. Intelligent Control Autom. WCICA}, 2012, pp. 4532--4535.

\bibitem{6706382}
H.-B. Zhang \emph{et~al.}, ``Seeing actions through scene context,'' in \emph{IEEE VCIP - IEEE Int. Conf. Vis. Commun. Image Process.}, 2013, pp. 1--6.

\bibitem{6977062}
J.~Liu, X.~Wu, and Y.~Feng, ``Modeling the relationship of action, object, and scene,'' in \emph{Proc. Int. Conf. Pattern Recognit.}, 2014, pp. 2005--2010.

\bibitem{liu2009recognizing}
J.~Liu, J.~Luo, and M.~Shah, ``Recognizing realistic actions from videos “in the wild”,'' in \emph{Proc. IEEE Conf. Comput. Vis. Pattern Recog.}\hskip 1em plus 0.5em minus 0.4em\relax IEEE, 2009, pp. 1996--2003.

\bibitem{reddy2013recognizing}
K.~K. Reddy and M.~Shah, ``Recognizing 50 human action categories of web videos,'' \emph{Mach. Vision Appl.}, vol.~24, no.~5, pp. 971--981, 2013.

\bibitem{7532632}
H.~Wang, W.~Wang, and L.~Wang, ``How scenes imply actions in realistic videos?'' in \emph{Proc. Int. Conf. Image Process. ICIP}, 2016, pp. 1619--1623.

\bibitem{8101020}
J.~Hou, X.~Wu, Y.~Sun, and Y.~Jia, ``Content-attention representation by factorized action-scene network for action recognition,'' \emph{IEEE Trans. Multimedia}, vol.~20, no.~6, pp. 1537--1547, 2018.

\bibitem{10006458}
J.~He, X.~Zhao, B.~Sun, X.~Yu, and Y.~Zhang, ``Visual scene induced three-stream network for efficient action recognition,'' in \emph{ICICN - IEEE Int. Conf. Inf., Commun. Networks}, 2022, pp. 550--554.

\bibitem{10378184}
Y.~Zhai \emph{et~al.}, ``Soar: Scene-debiasing open-set action recognition,'' in \emph{Proc. IEEE Int. Conf. Comput. Vis.}, 2023, pp. 10\,210--10\,220.

\bibitem{wang2020eca}
Q.~Wang \emph{et~al.}, ``Eca-net: Efficient channel attention for deep convolutional neural networks,'' in \emph{Proc. IEEE Conf. Comput. Vis. Pattern Recog.}, 2020, pp. 11\,534--11\,542.

\bibitem{10007033}
Y.~Zhao \emph{et~al.}, ``A novel action saliency and context-aware network for weakly-supervised temporal action localization,'' \emph{IEEE Trans. Multimedia}, vol.~25, pp. 8253--8266, 2023.

\bibitem{https://doi.org/10.48550/arxiv.2302.12288}
\BIBentryALTinterwordspacing
S.~F. Bhat, R.~Birkl, D.~Wofk, P.~Wonka, and M.~M{\"u}ller, ``Zoedepth: Zero-shot transfer by combining relative and metric depth,'' 2023. [Online]. Available: \url{https://arxiv.org/abs/2302.12288}
\BIBentrySTDinterwordspacing

\bibitem{liang2023open}
F.~Liang \emph{et~al.}, ``Open-vocabulary semantic segmentation with mask-adapted clip,'' in \emph{Proc. IEEE Conf. Comput. Vis. Pattern Recog.}, 2023, pp. 7061--7070.

\bibitem{sun2023siamohot}
C.~Sun \emph{et~al.}, ``Siamohot: A lightweight dual siamese network for onboard hyperspectral object tracking via joint spatial-spectral knowledge distillation,'' \emph{IEEE Trans. Geosci. Remote Sens.}, 2023.

\bibitem{shen2023indoor}
M.~Shen, K.-L. Tsui, M.~A. Nussbaum, S.~Kim, and F.~Lure, ``An indoor fall monitoring system: Robust, multistatic radar sensing and explainable, feature-resonated deep neural network,'' \emph{IEEE J. Biomed. Health. Inf.}, vol.~27, no.~4, pp. 1891--1902, 2023.

\bibitem{song2024learning}
Y.~Song, S.~Xia, J.~Yang, and L.~Pei, ``A learning-based multi-node fusion positioning method using wearable inertial sensors,'' in \emph{ICASSP IEEE Int Conf Acoust Speech Signal Process Proc}.\hskip 1em plus 0.5em minus 0.4em\relax IEEE, 2024, pp. 1976--1980.

\bibitem{bertasius2021space}
G.~Bertasius, H.~Wang, and L.~Torresani, ``Is space-time attention all you need for video understanding?'' in \emph{ICML}, vol.~2, no.~3, 2021, p.~4.

\bibitem{liu2022video}
Z.~Liu \emph{et~al.}, ``Video swin transformer,'' in \emph{Proc. IEEE Conf. Comput. Vis. Pattern Recog.}, 2022, pp. 3202--3211.

\bibitem{li2022mvitv2}
Y.~Li \emph{et~al.}, ``Mvitv2: Improved multiscale vision transformers for classification and detection,'' in \emph{Proc. IEEE Conf. Comput. Vis. Pattern Recog.}, 2022, pp. 4804--4814.

\bibitem{li2023uniformerv2}
K.~Li \emph{et~al.}, ``Uniformerv2: Unlocking the potential of image vits for video understanding,'' in \emph{Proc. IEEE Int. Conf. Comput. Vis.}, 2023, pp. 1632--1643.

\bibitem{yan2018spatial}
S.~Yan, Y.~Xiong, and D.~Lin, ``Spatial temporal graph convolutional networks for skeleton-based action recognition,'' in \emph{AAAI - AAAI Conf. Artif. Intell.}, vol.~32, no.~1, 2018.

\bibitem{shi2019two}
L.~Shi, Y.~Zhang, J.~Cheng, and H.~Lu, ``Two-stream adaptive graph convolutional networks for skeleton-based action recognition,'' in \emph{Proc. IEEE Conf. Comput. Vis. Pattern Recog.}, 2019, pp. 12\,026--12\,035.

\bibitem{duan2022pyskl}
H.~Duan, J.~Wang, K.~Chen, and D.~Lin, ``Pyskl: Towards good practices for skeleton action recognition,'' in \emph{MM - Proc. ACM Int. Conf. Multimed.}, 2022, pp. 7351--7354.

\bibitem{kondratyuk2021movinets}
D.~Kondratyuk \emph{et~al.}, ``Movinets: Mobile video networks for efficient video recognition,'' in \emph{Proc. IEEE Conf. Comput. Vis. Pattern Recog.}, 2021, pp. 16\,020--16\,030.

\end{thebibliography}

\end{document}